# Product safety idioms: a method for building causal Bayesian networks for product safety and risk assessment

Joshua L. Hunte[a]*, Martin Neil[a,b] and Norman E. Fenton[a,b]

[a]Risk and Information Management Research Group, School of Electronic Engineering and Computer Science, Queen Mary University of London, London, E1 4NS, UK; [b]Agena Ltd, Cambridge, UK

*Corresponding author address: Joshua Hunte, Risk and Information Management Research Group, School of Electronic Engineering and Computer Science, Queen Mary University of London, London, E1 4NS, UK. email address: j.l.hunte@qmul.ac.uk

**Abstract**

Idioms are small, reusable Bayesian network (BN) fragments that represent generic types of uncertain reasoning. This paper shows how idioms can be used to build causal BNs for product safety and risk assessment that use a combination of data and knowledge. We show that the specific product safety idioms that we introduce are sufficient to build full BN models to evaluate safety and risk for a wide range of products. The resulting models can be used by safety regulators and product manufacturers even when there are limited (or no) product testing data.

# 1. Introduction

Product safety and risk assessments are performed by safety regulators, product manufacturers, and safety and risk analysts to ensure that products or systems available on the market are sufficiently safe for use. There are several methods of product safety and risk assessment used in different domains; for instance, RAPEX (European Commission, 2018) is used for consumer products, Fault Tree Analysis (FTA), Event Tree Analysis (ETA) and Failure Mode and Effect Analysis (FMEA) are commonly used for many safety-critical products or systems, including medical devices and aerospace applications (Elahi, 2022; ISO, 2020; SAE, 2012; Vesely, W. E., Dugan, J., Fragola, J., Minarick, J., & Railsback, 2002).

Special challenges for these methods include being able to handle: dependencies among system components (including common causes of errors and failures); full quantification of uncertainty; limited relevant testing and failure data for novel products; and rigorous ways to incorporate expert judgment (Fenton & Neil, 2018; Hunte, Neil, & Fenton, 2022; Kabir & Papadopoulos, 2019; Weber, Medina-Oliva, Simon, & Iung, 2012). Bayesian networks are able to address all of the challenges (Berchialla et al., 2016; Fenton & Neil, 2018; Hunte et al., 2022; Kabir & Papadopoulos, 2019; Ruijters & Stoelinga, 2015; Suh, 2017). For instance, Hunte et al. (2022) (as part of work in collaboration with the UK Office for Product Safety and Standards) resolved the limitations of RAPEX using a Bayesian network (BN) for consumer product safety and risk assessment. Their proposed BN has shown good predictive performance and – unlike RAPEX – provides auditable quantitative safety and risk assessments even when there is little or no product testing data by combining objective and subjective evidence, i.e., data and knowledge.

We believe that BNs are suitable for product safety and risk assessments generally since it is a normative, rigorous method for modelling risk and uncertainty that is increasingly being used for system safety, reliability and risk assessments in several domains such as medical, railway and maritime (Fenton & Neil, 2010, 2018; Hänninen, Valdez Banda, & Kujala, 2014; Kabir & Papadopoulos, 2019; Li, Liu, Li, & Liu, 2019; Marsh & Bearfield, 2004; Weber et al., 2012). Also, BNs extend and complement classical methods to improve safety and risk assessment results. For instance, fault trees (FTs) and FMEA have been translated to BNs, allowing them to handle uncertainty and perform predictive and diagnostic analysis (Bobbio, Portinale, Minichino, & Ciancamerla, 2001; Marsh & Bearfield, 2007; Martins & Maturana, 2013; Wang & Xie, 2004).

However, despite the benefits of using BNs for safety, reliability and risk assessments, their widespread use as a systematic method for product safety and risk assessment may have been impeded due to the lack of any standard method or guidelines for their development and validation for the many different types of product safety cases.

The most promising method for developing coherent BN models is to use an idioms based approach (Helsper & Van der Gaag, 2002; Koller & Pfeffer, 1997; Laskey & Mahoney, 1997; Neil, Fenton, & Nielsen, 2000). Idioms are reusable BN fragments representing common generic types of uncertain reasoning. Researchers have developed idioms for specific application domains such as legal reasoning (Fenton, Neil, & Lagnado, 2013; Lagnado, Fenton, & Neil, 2013) and medical decision making (Kyrimi et al., 2020). In this paper, we introduce a novel set of idioms, called *product safety idioms,* to enable a systematic method for developing BNs specifically for product safety and risk assessment. While the proposed idioms are sufficiently generic to be applied to a wide range of product safety cases, they are not prescriptive or complete and should be considered as a guide for developing suitable idioms for product safety and risk assessments given available product-related information.

The paper is organised as follows: Section 2 provides the relevant background material, namely the standard safety and risk assessment framework and definition of terms and overview of Bayesian networks. The full description of the novel product safety idioms is provided in Sections 3 and 4; in Section 3 are the idioms associated with the risk estimation phase; while Section 4 has the idioms associated with the risk evaluation phase. Complete examples of applying the method to product safety cases are presented in Section 5. Finally, our conclusions and recommendation for further work are presented in Section 6.

## 2. Background

### 2.1. Standard risk assessment framework and definition of terms

Since RAPEX is the primary method or guideline used for consumer product safety and risk assessment by safety regulators in the European Union (European Commission, 2018), in what follows, we use the RAPEX definitions and phases of the risk assessment process as shown in Figure 1.

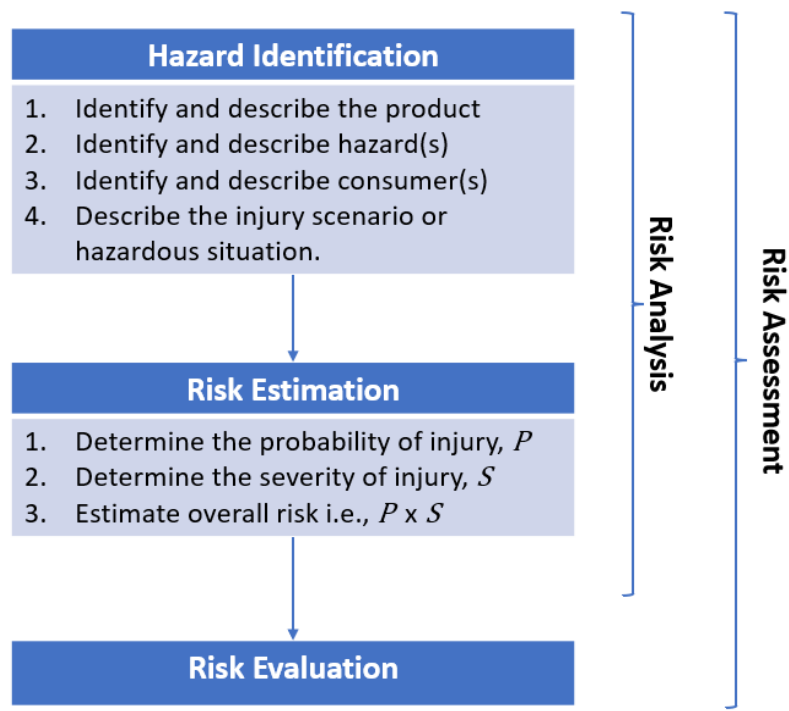

Figure 1. Overview of RAPEX risk assessment process

We use the terms *product* and *system* interchangeably. A *system* is a combination of interacting elements or components organised to achieve one or more stated purposes. Components of a system include hardware, software, material, facilities, personnel, data and services. A *product* is any artefact offered in a market to satisfy consumer needs. Hence, a product is a system, and a system can be described as a product or as the services it provides, such as the case of mobile phones (ISO/IEC/IEEE 15288, 2015).

The terms *defect*, *fault*, *error*, *failure* and *hazard* concerning a product, or a system are defined as follows: A *defect* is a generic term for a *fault*. A *fault* is a hypothesised cause of an error. An *error* is the part of the system state that can lead to a failure. A *failure* is an event that occurs when the delivered service deviates from fulfilling the system function (Laprie, 1995). A *hazard* is a potential source of harm, such as fire which can cause physical injury or damage to property (ISO, 2019). It is important to note that faults, errors, failures and hazards are recursive notions that depend on the perspective of the user and/or system. For example, for a system containing an embedded software component, a failure of the software component may not necessarily lead to a system failure - and hence would be classified as a fault from an overall system perspective.

The relationship between faults, errors, failures and hazards is shown in Figure 2. The three main types of faults associated with a system are physical faults, design faults and interaction faults (Laprie, 1995).

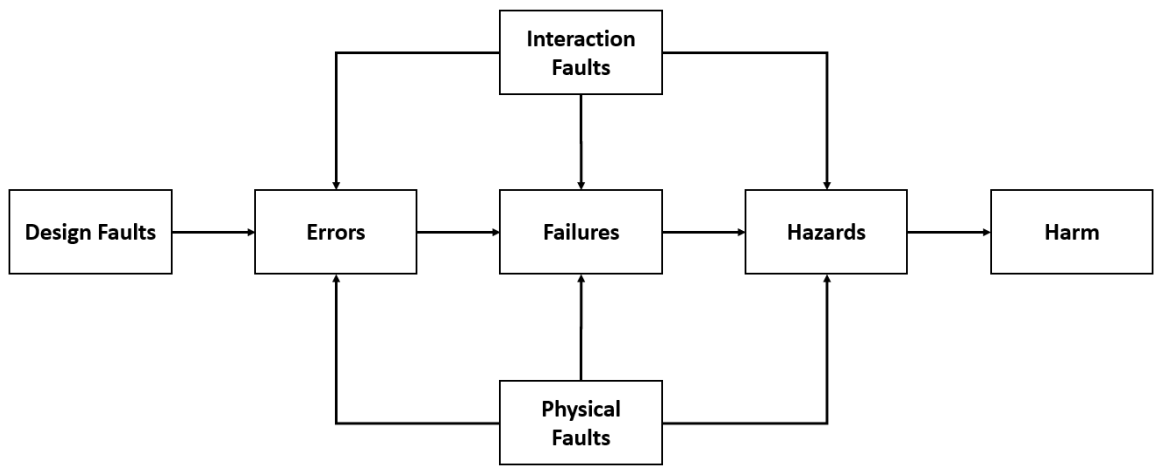

Figure 2. Relationship between system faults, errors, failures and hazards

Physical faults are faults in the hardware of a system (Avižienis, Laprie, Randell, & Landwehr, 2004; Laprie, 1995). They are caused by hardware deterioration, interaction faults and development faults, e.g., production defects. As shown in Figure 2, physical faults can cause (a) an error, (b) a failure in the absence of an error, (c) hazards in the absence of a failure.

Design faults are faults in the software of a system (Laprie, 1995). They are caused by interaction faults and development faults, e.g., errors in the design specification. As shown in Figure 2, design faults can cause a system error leading to failure and potential hazards.

Interaction faults are faults due to operational use or misuse of a system (Avižienis et al., 2004). These are external faults since they are caused by elements in the use environment, e.g., users. They include input mistakes and physical interference. As shown in Figure 2, interaction faults can cause (a) an error, (b) a failure in the absence of an error, (c) hazards in the absence of a failure.

As preparation for identifying the product safety idioms in Section 3, the following tasks are performed during the hazard identification phase (see Figure 1):

1. The product is identified. Information such as product name/type and model number are documented. For example, Product: Hammer, Model Number: 999.

2. All known and foreseeable hazards, e.g., hammer head detaching, associated with the product are identified and documented. Techniques used to identify potential faults and associated hazards include Preliminary Hazard Analysis (PHA), Fault Tree Analysis (FTA) and Failure Mode and Effect Analysis (FMEA) (ISO, 2020; SAE, 2012).

3. The consumers are identified. Since consumers' abilities and behaviour can affect overall product risk, information such as intended and non-intended users, duration and frequency of use are documented.

4. The injury scenario(s) is described. The injury scenario describes the steps to harm and usually consists of three main parts (a) the product has a fault that can cause a failure or hazard, (b) the failure or hazard leads to a hazardous situation, (c) the hazardous situation results in an injury. A *hazardous situation* is any circumstance where the user, property or environment is exposed to one or more product hazards (ISO, 2019). For example:
    a. *Hammer example*: The hammer head has been made from unsuitable material, and metal parts may detach or break and injure the person using the hammer or people nearby.

    b. *Car example*: The defective heat treatment of the engine components may lead to the failure of the engine, causing injuries.

The specific tasks associated with the risk estimation and risk evaluation phases of RAPEX will be defined as part of the idioms in Sections 3 and 4, respectively. There are no product safety idioms associated with the initial hazard identification phase that was described in Section 2.1; however, the information gathered in this phase is essential for identifying relevant variables affecting product risk. These identified variables are organised into idioms for risk estimation and risk evaluation.

## 2.2. Bayesian Networks

Bayesian networks (BNs) are probabilistic graphical models that describe the causal relationship between a set of random variables. A BN consists of two components: a directed acyclic graph (DAG) and (2) node probability tables (NPT) for each node in the BN (Fenton

& Neil, 2018; Pearl, 2009; Spohn, 2008). The DAG consists of a set of nodes and directed arcs. The nodes represent the random variables, and the directed arcs represent the causal relationship or causal influence between the nodes. For example, given two nodes, *A* and *B,* as shown in Figure 3, a directed arc from *A* to *B* indicates that *A* causally influences *B* or *B* is dependent on *A*; thus, *A* is called the parent of *B* and *B* is called the child of *A* (Fenton & Neil, 2018; Pearl & Mackenzie, 2018).

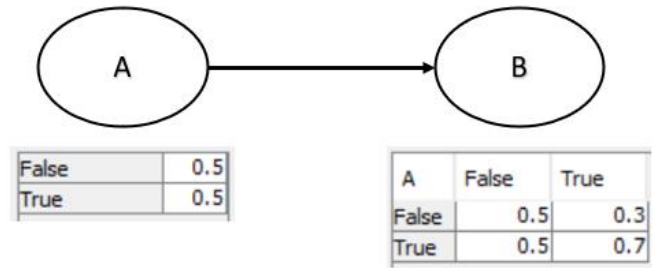

Figure 3. Two-node Bayesian Network

Each node in the DAG has a node probability table (NPT), as shown in Figure 3, that describes the probability distribution of the node conditioned on their parents. The particular functions and operators used to define the NPTs depend on the node type, i.e., discrete or continuous. For instance, the NPTs for discrete nodes are defined using functions and comparative expressions, e.g., NoisyOR, or manually as shown in Figure 3. The NPTs for continuous nodes are defined using conditionally deterministic functions, e.g., $C = A + B,$ and statistical distributions, e.g., Normal, Binomial and Exponential distributions. Any nodes without parents are called root nodes, and the NPTs for these nodes are their prior probability distributions (Fenton & Neil, 2018). Once all the NPTs are specified, the BN is fully parameterized and can be used to perform different types of probabilistic reasoning using Bayes Theorem. Bayes Theorem revises the initial belief of a given hypothesis when new evidence is available. The initial belief is called the prior (or prior probability), and the revised belief is called the posterior (or posterior probability) (Pearl, 2009).

The three main types of reasoning done using BNs are observational, interventional and counterfactual reasonings (Pearl, 2009; Pearl & Mackenzie, 2018). Observational reasoning entails entering an observation in the BN to discover its cause (i.e., diagnostic or backward inference) or to discover its effect (predictive or forward inference). Interventional reasoning entails fixing the value of a variable (also called intervening on the variable) to determine its effect on dependent (or child) variables. Intervening on the variable is done by removing the directed arcs between the node and its parents. Counterfactual reasoning entails using the BN to predict what would have happened if other events instead of the observed events had happened (Pearl, 2009; Pearl & Mackenzie, 2018). It can be performed using a twin network model containing two identical networks, one representing the real world and one representing the counterfactual world connected via shared background (exogenous) variables. In the twin network model, an observation in the real world is modelled as an intervention in the counterfactual world (Balke & Pearl, 1994; Pearl, 2009). The fact that BNs that represent casual knowledge can be used for both interventional and counterfactual reasoning rather than just observational reasoning is what makes them so powerful in comparison to typical statistical regression modelling and other machine learning methods (Pearl & Mackenzie, 2018).

## 3. Risk Estimation idioms

The second phase of RAPEX's risk assessment process is risk estimation (see Figure 1). Given the injury scenario for a product described during the hazard identification phase, the risk of a product is computed as $P \times S$, where $P$ is the probability of injury and $S$ is the severity of the injury. The probability of injury $P$ is estimated by assigning probabilities to each step of the injury scenario and multiplying them together. Therefore, to estimate $P$, the risk assessor needs to determine the (1) probability of failure or hazard (2) probability of failure or hazard leading to a hazardous situation (3) probability of harm given the hazardous situation. The severity of the injury $S$ is determined by the level of medical intervention required for the injury described in the injury scenario. It ranges from levels 1 to 4, where level 1 indicates an injury requiring first aid treatment and level 4 indicates a fatal injury. The overall risk of the product is then determined by using a risk matrix that combines the probability of injury $P$ and the severity of injury $S$.

As was made clear in (Hunte et al., 2022) this method of estimating risk has several limitations which are resolved using BNs. In this section, we show how the information gathered during the hazard identification phase can be organised into novel idioms that can be combined and reused as required to arrive at a rigorous, systematic, and properly quantified estimate of the overall risk of a product. Since this entails determining the occurrence of failures or hazards and related injuries, the idioms are grouped based on their scope as follows:

1. *Reliability*: These idioms are used to estimate the reliability of a product in terms of failure rate (i.e., probability of failure on demand and time to failure). They model the results of product testing to estimate the probability of a failure or hazard occurring for a product.

2. *Rework or Maintenance*: These idioms are used to model the probability of repairing identified faults of a product.

3. *Requirement*: These idioms are used to determine whether the product satisfies defined operational and safety requirements.

4. *Quality*: These idioms are used to estimate the quality of a particular entity or process that may affect the overall reliability of a product, e.g., manufacturing process quality.

5. *Failure, Hazard and Injury Occurrence*: These idioms are used to estimate the hazard or failure occurrence and associated injuries for a product given factors such as consumer behaviour.

6. *Risk:* These idioms are used to estimate the overall risk level of the product.

We will base our discussion on the idioms by using the hammer and car injury scenario examples described in Section 2.1

### 3.1. Idioms for modelling reliability

Determining the reliability of a product is important for informing risk controls and rework activities since failures and hazards can cause harm to users and damage to the environment. Building a practical BN for a product safety case requires the risk assessor(s) to identify and understand the reliability or safety metric for that system. The two main reliability metrics for systems are probability of failure on demand (PFD) and time to failure (TTF) (Rausand & Hoyland, 2003). Probability of failure on demand (PFD) relates to the reliability associated with a finite set of uses of the system. For instance, if the system is a car, then we might be interested in the probability of failure for a given type of journey. In contrast, time to failure (TTF) relates to reliability associated with a system operating in continuous time. For instance, for a car, we may also be interested in the number of miles it could drive before a failure occurs. For complex systems such as an aircraft, it is inevitable that we will need to consider both TTF and PFD measures to determine its overall reliability because some of its sub-systems like the engine require the TTF measure while others like the landing gear system require the PFD measure.

In Subsection 3.1.1, we describe idioms associated with determining PFD and in Subsection 3.1.2, we describe idioms associated with determining TTF.

#### 3.1.1. Idioms for Probability of failure on demand (PFD)

There are three idioms in this category:

1. Hazard or failure per demand idiom (generic)
2. Hazard or failure per demand with limited data idiom
3. Probability of an event with uncertain accuracy idiom

Please note that the proposed idioms for handling limited data and uncertain accuracy are situational; model experts may develop other idioms based on the type of censored data.

**Hazard or failure per demand idiom (generic)**

During the hazard identification phase, techniques such as Preliminary Hazard Analysis (PHA) and Failure Mode and Effect Analysis (FMEA) are used to identify potential hazards and failures for a product. Once hazards and failures are identified, product testing is done to quantify and learn the 'true' reliability or safety of the product. Product testing entails observing the product being used many times and recording each observed failure or hazard, respectively. We define a *demand* as a measure of usage; for example, a washing machine is typically used on average 200 times per year in each UK home that has one. Some products, such as certain medical devices, are intended to be only used once, i.e., single-use devices. By observing a large number of demands of a product or product type and recording the number of demands which result in a hazard or failure, we can learn an estimate of the 'true' probability of hazard or failure per demand as a probability distribution. The more demands we observe, the smaller the variance (uncertainty) we have about this distribution.

The generic hazard or failure per demand idiom (shown in Figure 4a) models the probability distribution of the hazard or failure per demand based on the number of hazards or failures observed during a set of demands (trials). As shown in Table 1, this idiom uses a Binomial distribution for the number of times the hazard is observed. If there are no relevant prior data, it uses an 'ignorant' uniform prior for the probability of hazard (or failure) per demand. For instance, assuming a uniform prior for the hammer example (see Section 2.1), if we observe the hammer head detaching (hazard) 10 times in 1000 demands during testing, we can use this information to estimate the reliability of the hammer as a probability distribution. In Figure 4b, the idiom estimates that the mean probability of the hammer head detaching per demand is 0.01 with a variance of 1.11E-5.

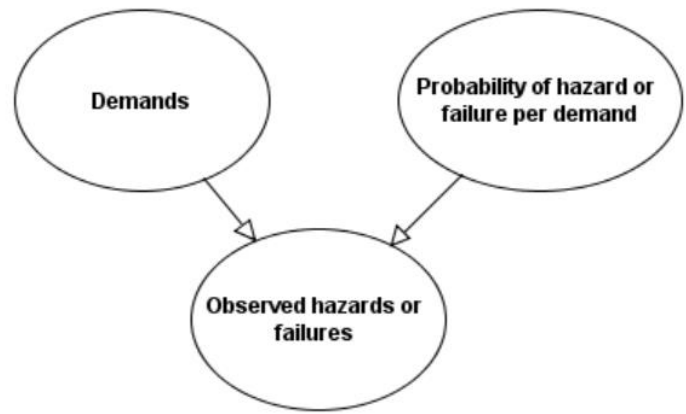

Figure 4a Hazard or failure per demand idiom (generic)

Table 1. NPTs for the nodes of the *Hazard or failure per demand idiom*

| Node Name | NPT |
| --- | --- |
| **Observed hazards or failures** | Binomial ($n$, $p$), where $n$ = demands and $p$ = probability of hazard or failure per demand |
| **Demands** | Uniform (0, 1E9) |
| **Probability of hazard or failure per demand** | Uniform (0,1) |

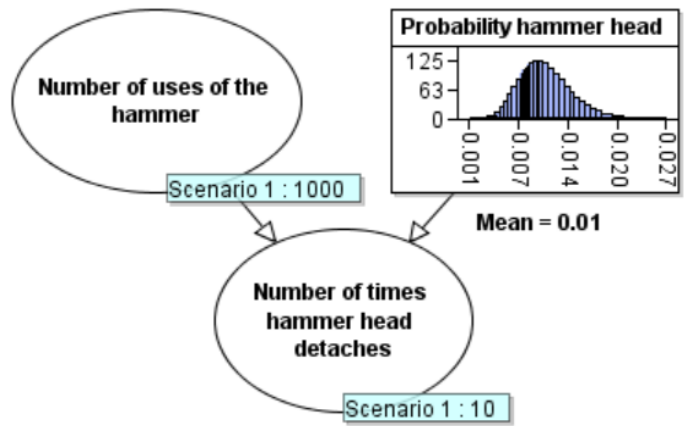

Figure 4b. Hazard or failure per demand idiom instance

**Hazard or Failure per demand with limited data idiom**

For some products, it will neither be feasible nor possible to get any extensive data from testing to estimate their 'true' reliability. In these situations, we can adapt the hazard or failure per demand idiom to incorporate testing data from previous similar products (if available) to estimate the 'true' reliability or safety of the product.

The hazard or failure per demand with limited data idiom is shown in Figure 5a, and instances are shown in Figures 5b and 5c. The NPT values for the node 'Probability of failure or hazard per demand' (see Table 2) can easily be adapted given the product. In Figure 5b, for the hammer example (see Section 2.1), we show that if we do not have any testing data for the hammer, we can estimate the reliability of the hammer using testing data from a previous similar hammer (200 failures in 2000 demands in this example). Given previous similar hammer data, the idiom estimates that the mean probability of the hammer head detaching (hazard) per demand is 0.125 with a variance of 1.9E-4. We assume that there were "minor differences" in the previous type of hammer and in its testing. In Figure 5c, given limited testing data for the hammer (0 hazards or failures in 500 demands in this example) together with testing data from a previous similar hammer (200 failures in 2000 demands in this example), the idiom estimates that the mean probability of hammer head detaching (hazard) per demand is 0.04 with a variance of 2.7E-4.

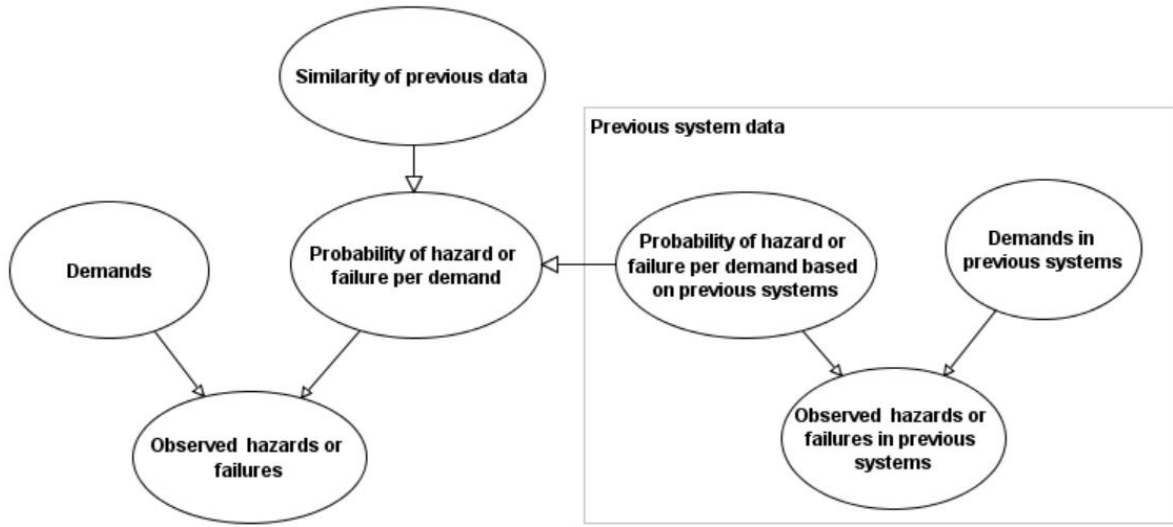

Figure 5a. Hazard or failure per demand with limited data idiom

Table 2. NPT for the node *Probability of hazard or failure per demand*

| Parent (Similarity of previous data) states | Probability of hazard or failure per demand |
|---|---|
| **Similar** | Normal (*pfd*, 1E-4), where *pfd* = probability of hazard per demand for the previous system |
| **Minor differences** | Normal (*pfd* × 1.25, 1E-4) |
| **Major differences** | Normal (*pfd* × 2, 1E-4) |

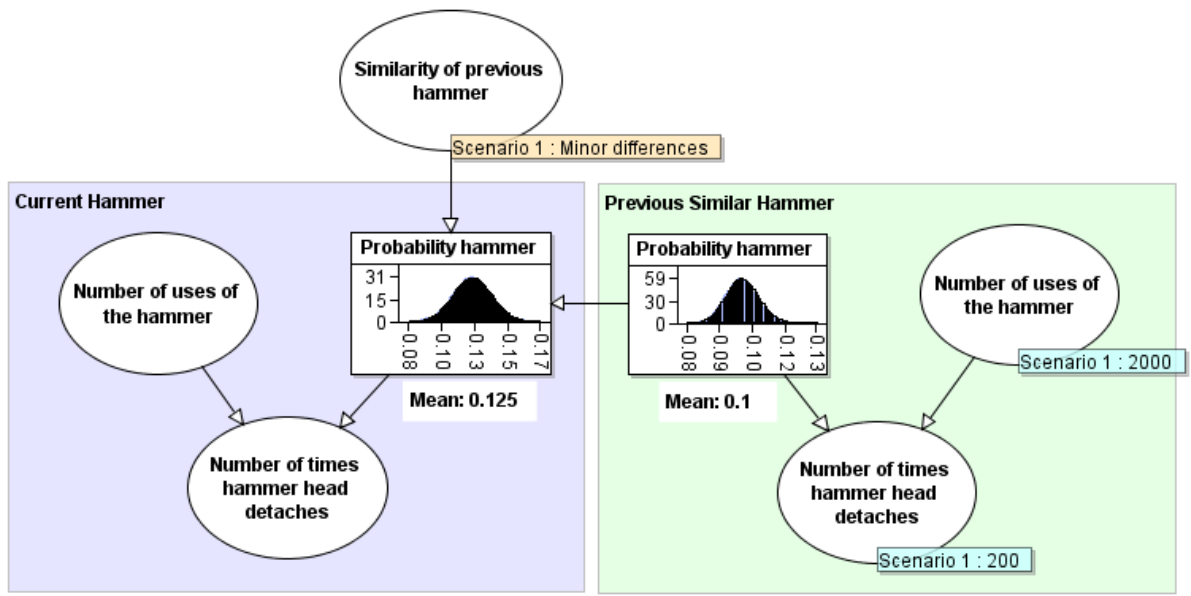

Figure 5b. Hazard or failure per demand with limited data idiom instance 1

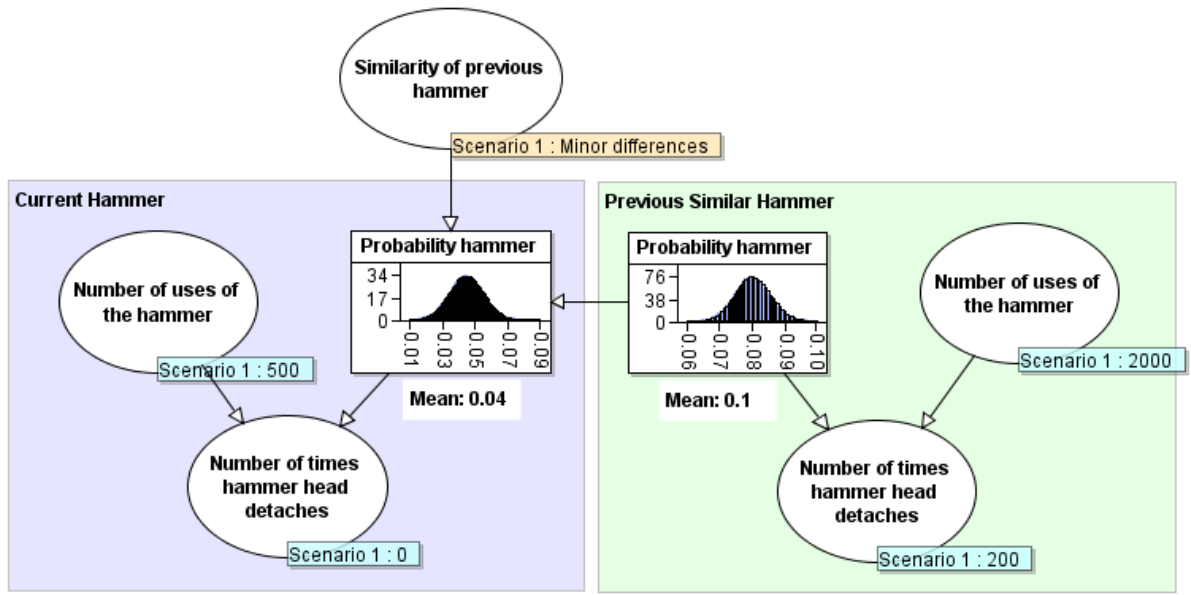

Figure 5c. Hazard or failure per demand with limited data idiom instance 2

**Probability of an event with uncertain accuracy idiom**

For some products, there may be uncertainty concerning the number of observed hazards or failures and, subsequently their 'true' reliability or safety. In these situations, we need to consider the accuracy of the number of observed hazards or failures and the true number of observed hazards or failures given our knowledge about the former when estimating the 'true' reliability of the product.

The probability of an event with uncertain accuracy idiom shown in Figure 6a models the uncertainty concerning the number of observed events e.g., hazards, failures or injuries (it can also be adapted to model the uncertainty concerning the number of trials or demands). The NPT values for the node 'Number of observed events' (see Table 3) can easily be adapted given the product. In Figure 6b, for the hammer example (see Section 3.1), if we assume that the number of times we observe the hammer head detaching (100 in this example) given a set of demands (1000 here) is underestimated, then the true number of times the hammer head detaches will be greater than the number of times we observed (in this example the mean of the true number of times the hammer head detaches is 125).

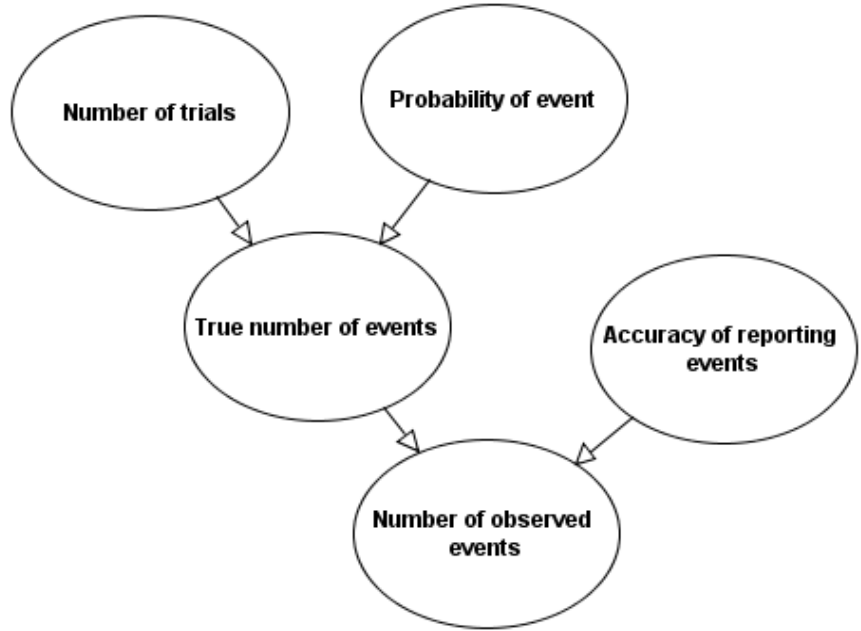

Figure 6a. Probability of an event with uncertain accuracy idiom

Table 3. NPT for the node *Number of observed events*

| Parent (Accuracy of reporting events) states | Number of observed events |
|---|---|
| **Overestimated** | Normal ($tne \times 1.2$, $1E-4 \times tne$), where $tne$ = true number of events |
| **Accurate** | Arithmetic(*tne*) |
| **Underestimated** | Normal (max (0, $tne \times 0.8$), $1E-4 \times tne$) |

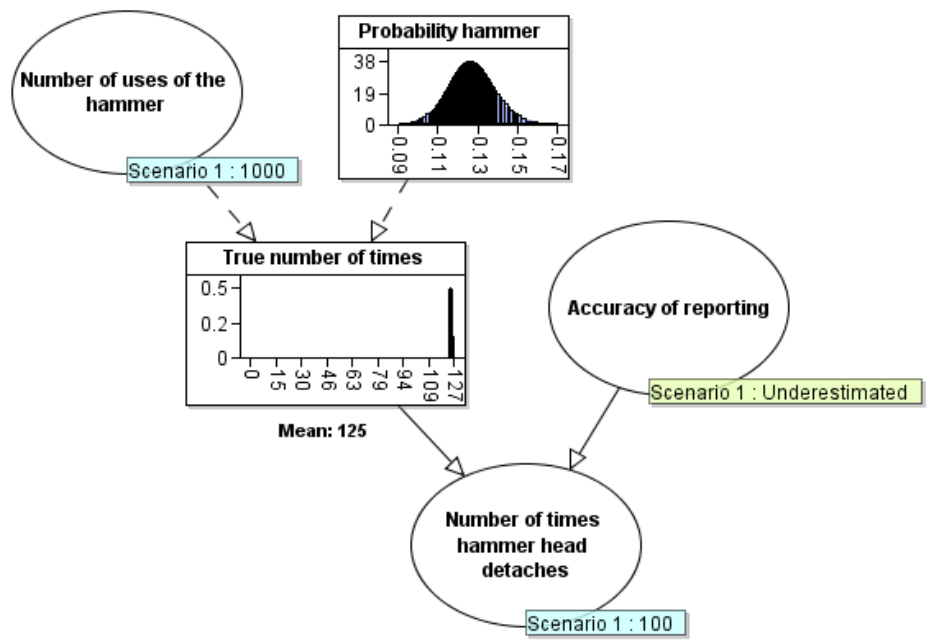

Figure 6b. Probability of an event with uncertain accuracy idiom instance

### 3.1.2. Idioms for Time to Failure

There are three idioms in this category:
1. Time to failure (or hazard) idiom (generic)
2. Time to failure (or hazard) idiom with summary statistics
3. Probability of failure within a specified time idiom

**Time to failure (or hazard) idiom (generic)**

For some products, we are interested in the reliability associated with the product operating in continuous time. In these situations, we can estimate the mean time to (next) failure by learning the time to failure (TTF) distribution of the product using failure data from testing or operational field use. The mean time to (next) failure is the summary statistic of the time to failure (TTF) distribution. The failure data will be a unit of time such as hours and may come from previous similar products. However, please note that model experts may develop other TTF idioms to estimate reliability given available TTF data and other related issues such as censoring.

The time to failure idiom shown in Figure 7a estimates the mean time to (next) failure for a product when there is a small number *n* of observed failure times. This idiom has *n* observed failure time nodes, which are used to estimate the failure rate of the product. The 'Observed failure time' and 'Time to next failure' nodes are (normally) defined as an Exponential distribution with the rate parameter as the value of the 'Assessed failure rate' node. Other distributions such as Weibull and Gamma can be used to define the nodes since the failure rate for many products is not usually constant but increases with time due to system use. However, please note that for the TTF idioms discussed in this paper we are assuming neither system improvement nor degradation and hence the time to (next) failure is constant. An instance of this idiom is shown in Figure 7b. In Figure 7b, for the car example (see Section 2.1), the TTF idiom estimates that the mean time to (next) failure for the car engine is 100 and the failure rate is 0.01 given observed failure times of 80, 90, 110 and 120, respectively.

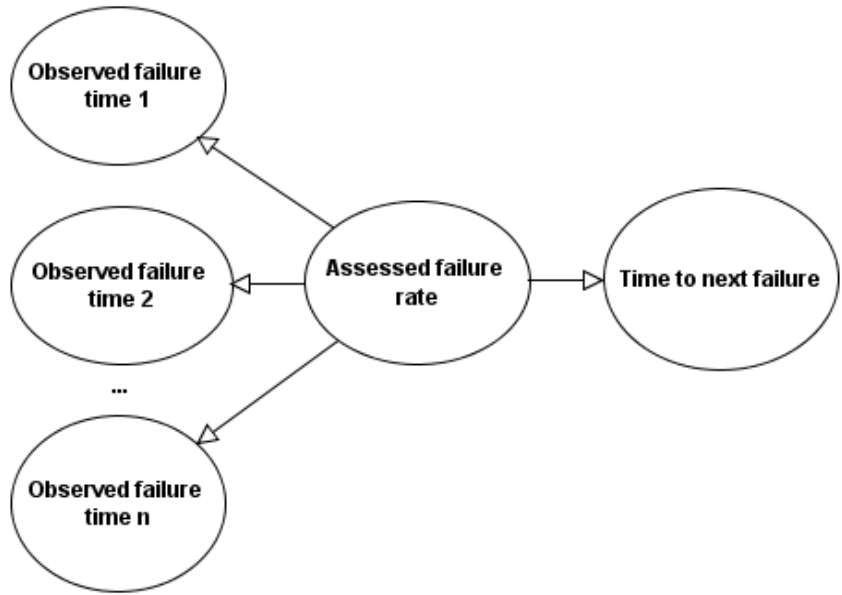

Figure 7a. Time to failure (or hazard) idiom

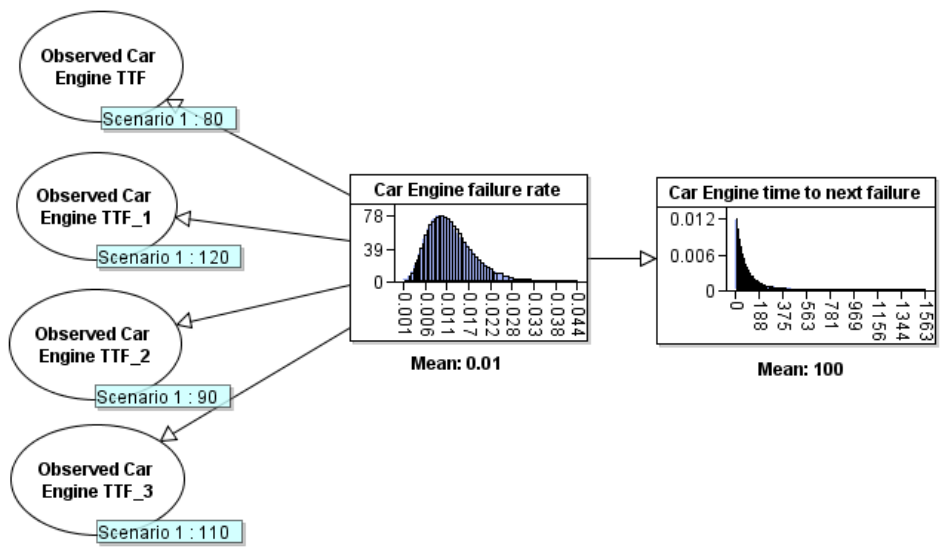

Figure 7b. Time to failure (or hazard) idiom instance

**Time to failure (or hazard) idiom with summary statistics**

For some products, there may be a large number of observed failures times. In these situations, it is more convenient to summarise the observed failure times in terms of their mean $\mu$ and variance $\sigma^2$ and use these as parameters to determine the rate value (i.e., $\frac{1}{Observed\ failure\ time}$) of an Exponential distribution. However, please note that this approach for handling a large number of observed failure times is situational, and the results are less accurate than using the generic TTF idiom; model experts may develop other TTF idioms to estimate reliability given available TTF data and other related issues such as censoring.

The time to failure idiom with summary statistics is shown in Figure 8a, and an instance is shown in Figure 8b. In Figure 8b, for the car example (see Section 2.1), the TTF idiom estimates that the mean time to (next) failure for the car engine is 100, given that the mean $\mu$ observed failure time is 100 and variance $\sigma^2$ is 250.

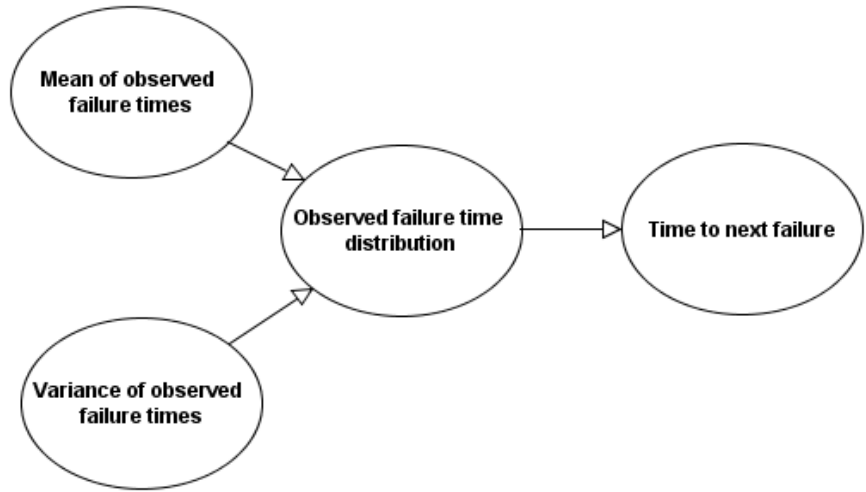

Figure 8a. Time to failure (or hazard) idiom with summary statistics

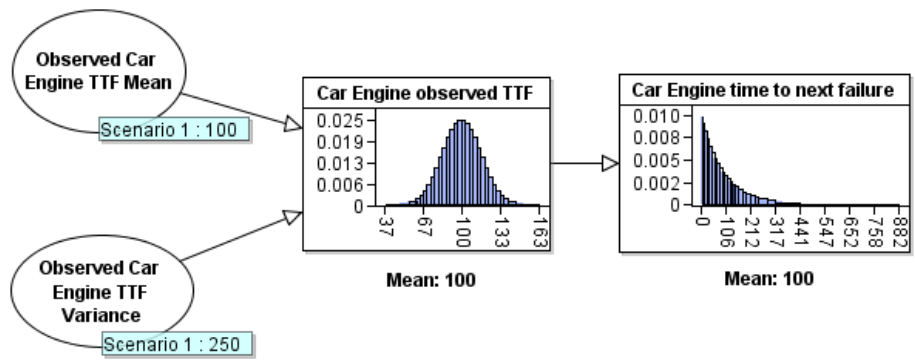

Figure 8b. Time to failure (or hazard) idiom with summary statistics instance

**Probability of failure within a specified time idiom**

For some products, we are interested in the reliability of the product operating within a specified time $t$. In these situations, we can estimate the probability of failure (or hazard) for a product within a specified time $P(Failure \mid t)$ by computing the probability that the TTF distribution $T$ is less than or equal to the specified time $t$, i.e., $P(Failure \mid t) = P(T \leq t)$. The probability of failure within a specified time idiom shown in Figure 9a uses a discrete node called 'Assessed probability of failure' to compute $P(T \leq t)$. The TTF distribution $T$ will be derived from the previous TTF idioms. An instance of this idiom is shown in Figure 9b. In Figure 9b, for the car example (see Section 2.1), the idiom estimates that if the car is used continuously for 10 hours, then the probability that the engine will fail is 0.1 or 10% given that the estimated mean time to next failure is 100.

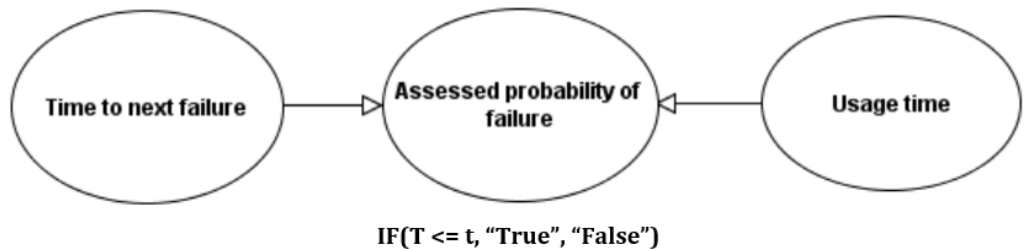

Figure 9a. Probability of failure within a specified time idiom

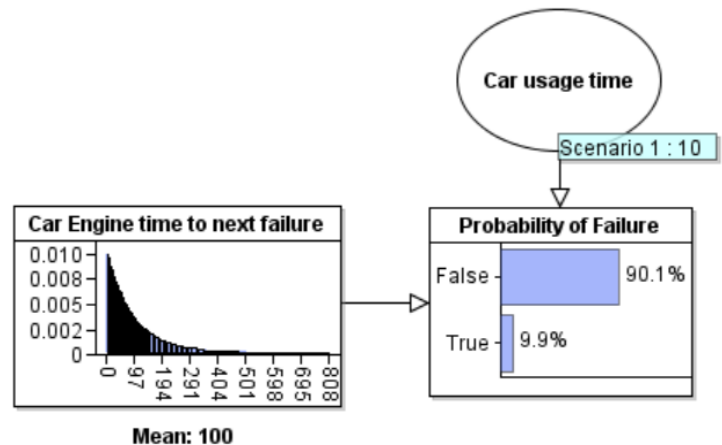

Figure 9b. Probability of failure within a specified time idiom instance

### 3.1.3. Rework idiom

For some products, faults identified during the hazard identification phase are repairable; however, the success of the repair will depend on the probability of fixing the fault. The rework idiom (Fenton, Neil, Marsh, et al., 2007) shown in Figure 10a incorporates knowledge of the manufacturer's rework process quality and rework effort to estimate the probability of fixing the fault (i.e., design and physical faults). This idiom uses ranked nodes (Fenton, Neil, & Caballero, 2007) to define 'rework process quality' and 'rework effort' since their values can be measured using a subjective ranked scale such as {'low', 'medium', 'high'}. These nodes are then combined to determine 'rework process overall effectiveness' (also a ranked node) and the 'probability of fixing the fault' (defined as a continuous node ranging from 0 to 1). The NPTs for the nodes in the idiom (see Table 4) can easily be adapted given the product. An instance of this idiom is shown in Figure 10b. In Figure 10b, for the hammer example (see Section 2.1), the idiom shows that if the manufacturer's rework process quality and effort are 'very low', then the overall rework process quality is also 'very low' or 'low'. As a result, the mean probability of fixing the hammer is very low (i.e., 0.03). Product manufacturers and safety regulators may use or adapt this idiom to revise the estimated reliability of the product given rework and to inform risk management decisions such as a product recall.

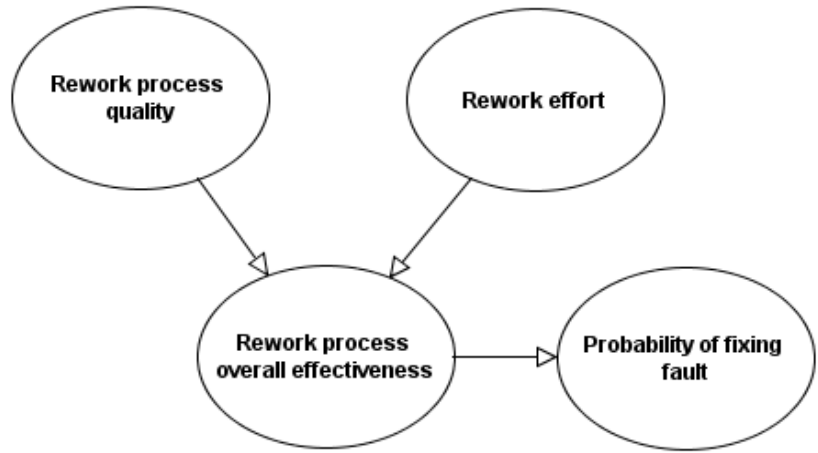

Figure 10a. Rework idiom

Table 4. NPTs for the nodes of the *Rework idiom*

| Node Name | NPT |
|---|---|
| **Rework process quality** | States ('very low', 'low', 'medium', 'high', 'very high') = 0.2 |
| **Rework effort** | States ('very low', 'low', 'medium', 'high', 'very high') = 0.2 |
| **Rework process overall effectiveness** | TNormal (wmean(1.0,rework_process,1.0,rework_effort), 0.001, 0, 1) |
| **Probability of fixing fault** | Partitioned expression ( <br> Very low: TNormal(0.01,0.001,0.0,1.0), <br> Low: TNormal(0.15,0.001,0.0,1.0), <br> Medium: TNormal(0.4,0.001,0.0,1.0), <br> High: TNormal(0.6,0.001,0.0,1.0), <br> Very High: TNormal(0.8,0.001,0.0,1.0)) |

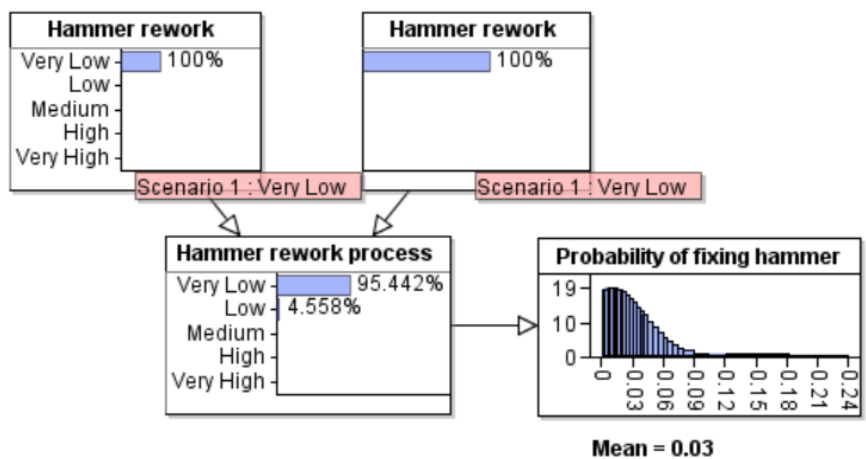

Figure 10b. Rework idiom instance

### 3.1.4. Requirement idiom

For any product, we will be interested in whether the safety and reliability of the product satisfy safety and reliability requirements defined by standards or safety regulators. Defined safety and reliability requirements ensure that a system operates as intended and is acceptably safe for use. For instance, as an extreme example, commercial aircraft must satisfy a defined safety and reliability requirement of MTTF $> 10^9$ flying hours to be approved for commercial use. Hence to determine if a product is compliant, we need to consider the defined safety and reliability value and the actual safety and reliability value of the product. However, testing alone may not be sufficient to determine the actual safety and reliability value of products, especially those with very high reliability requirements e.g., commercial aircraft or limited testing data e.g., novel products. In these situations, we need to combine testing information with other factors such as information about the quality of the processes and people involved in product development to determine the actual safety and reliability value of a product. The quality of processes or people can be estimated using the *Quality idiom* (see Section 3.1.5).

The requirement idiom shown in Figure 11a models whether the actual value of an attribute $A$ satisfies the defined requirement value of the attribute $R$ by computing the probability $A$ is less than or equal to $R$, i.e., $P(Compliant) = P(A \leq R)$. This idiom uses a discrete node called 'Assessed value of attribute' to compute $P(A \leq R)$. In the idiom instance shown in Figure 11a, for the hammer example (see Section 2.1), the idiom estimates that there is a 15% chance that the defined safety requirement (0.01 in this example) is satisfied given the probability distribution of the hammer head detaching (hazard) per demand (mean 0.03 in this example). Please note that the requirement idiom can also be implemented by encoding the requirement value into the 'Assessed value or attribute' node, as shown in Figure 11b. Product manufacturers and safety regulators may use or adapt the requirement idiom to inform risk management decisions such as a rework.

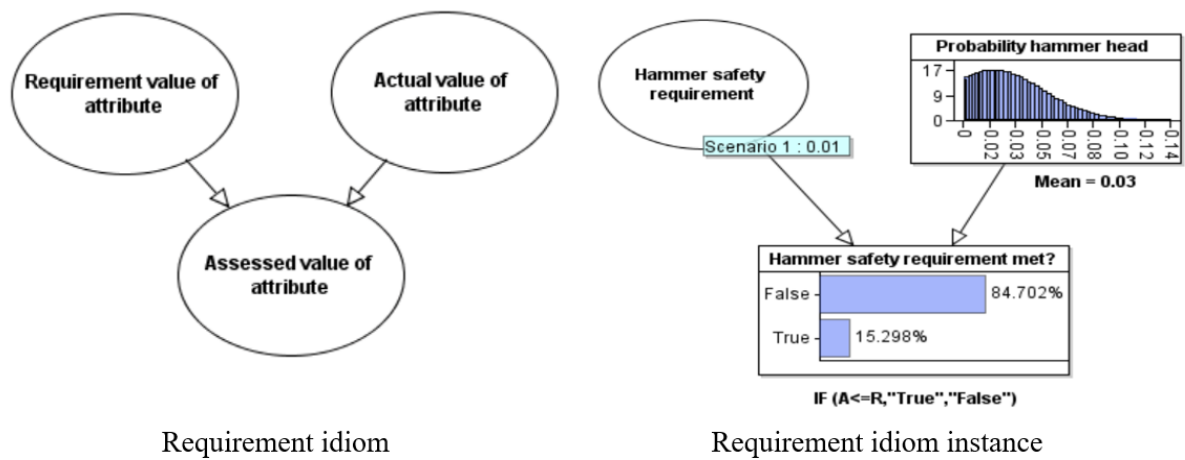

    Requirement idiom                  Requirement idiom instance

Figure 11a. Requirement idiom and instance

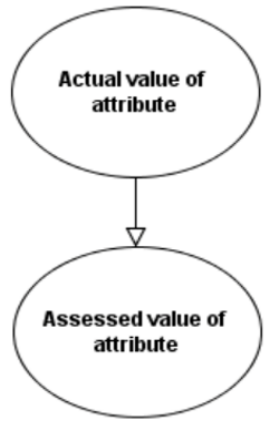 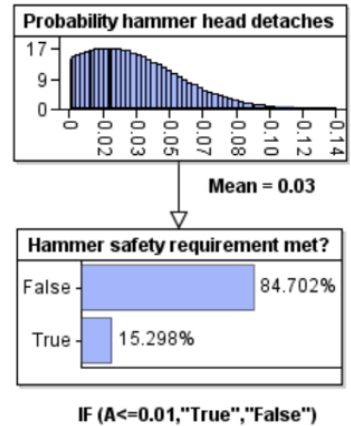

Implicit Requirement idiom         Implicit Requirement idiom instance

Figure 11b. Implicit Requirement idiom and instance

### 3.1.5. Quality idiom

For novel products, products with limited testing data and products with very high reliability requirements, other product-related information such as the quality of the processes and people involved in its development can be considered when estimating the reliability of the product. For instance, for the hammer example (see Section 2.1), if manufacturing process quality is poor, this can increase the likelihood of the hammer head detaching. However, the quality of a particular process or activity, such as the manufacturing process, may be latent, difficult to measure or observe. In these situations, we can use measurable indicators and causal factors to measure the quality of a particular process or activity.

The quality idiom (shown in Figure 12a) models the quality of an activity, process or variable using indicators and causal factors. This idiom uses ranked nodes (Fenton, Neil, & Caballero, 2007) to define variables since their values can be measured using a subjective ranked scale such as {'low', 'medium', 'high'}. Please note that NPT values for the node 'Latent quality value' (see Figure 12a) can easily be adapted given the process or activity. Instances of this idiom are shown in Figure 12b and Figure 12c for the hammer example. In Figure 12b, the idiom measures the quality of the manufacturing process, using knowledge about product defects and process drifts. In Figure 12c, the idiom measures the quality of the organisation using knowledge about customer satisfaction and years in operation.

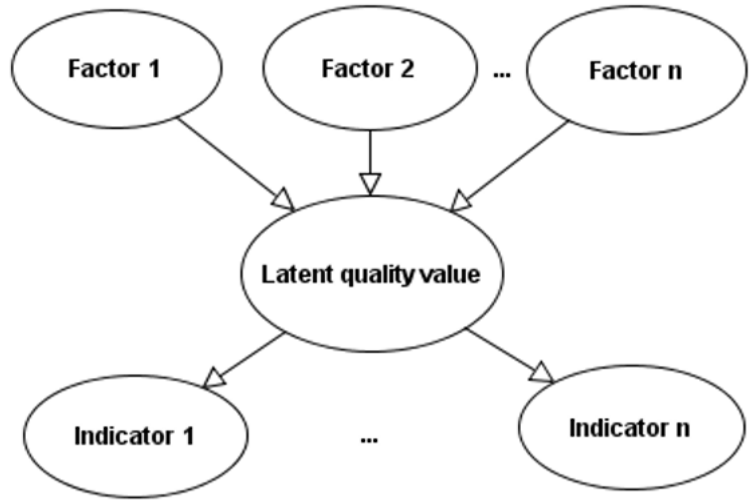

NPT for *Latent quality value* node: TNormal (wmean(1.0, Factor1, 1.0,Factor2, 1.0, Factorn), 0.001, 0, 1)

Figure 12a. Quality idiom

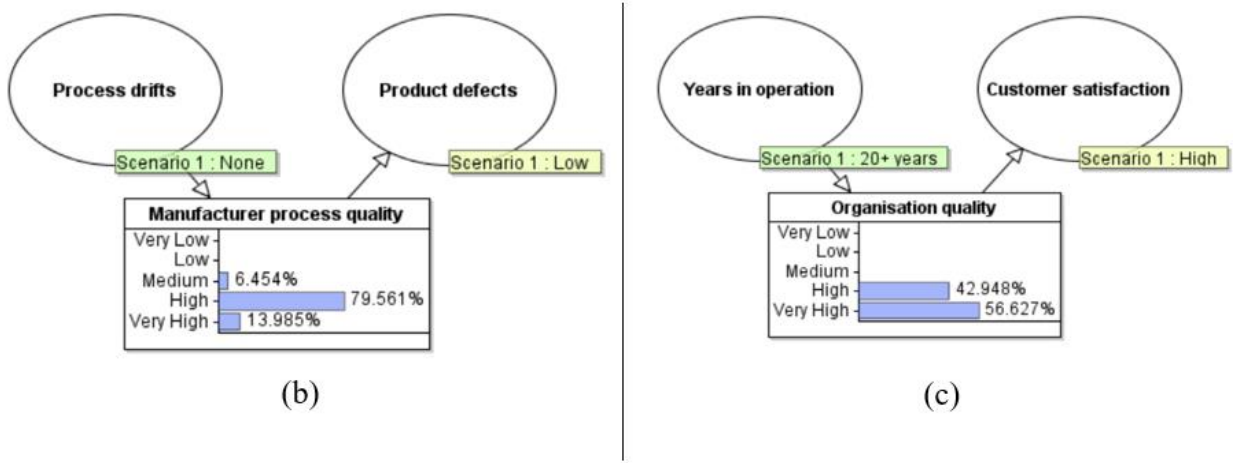

Figure 12 (b) Manufacturer process quality instance (c) Organisation quality instance

### 3.1.6. Combining product safety idioms to estimate product reliability

In Figure 13, for the hammer example (see Section 2.1), we show how the previously discussed idioms may be combined to determine the overall reliability of the hammer. In this example, using testing data only (i.e., hammer head detaches 20 times in 200 demands), the BN model estimates that the mean probability of the hazard per demand is 0.10. However, given information about the manufacturing process quality, the mean probability of the hazard per demand is revised. In this example, the mean probability of the hazard per demand increased to 0.15 due to a poor manufacturing process. Finally, the BN model shows that the reliability of the hammer did not satisfy the defined safety and operational requirement.

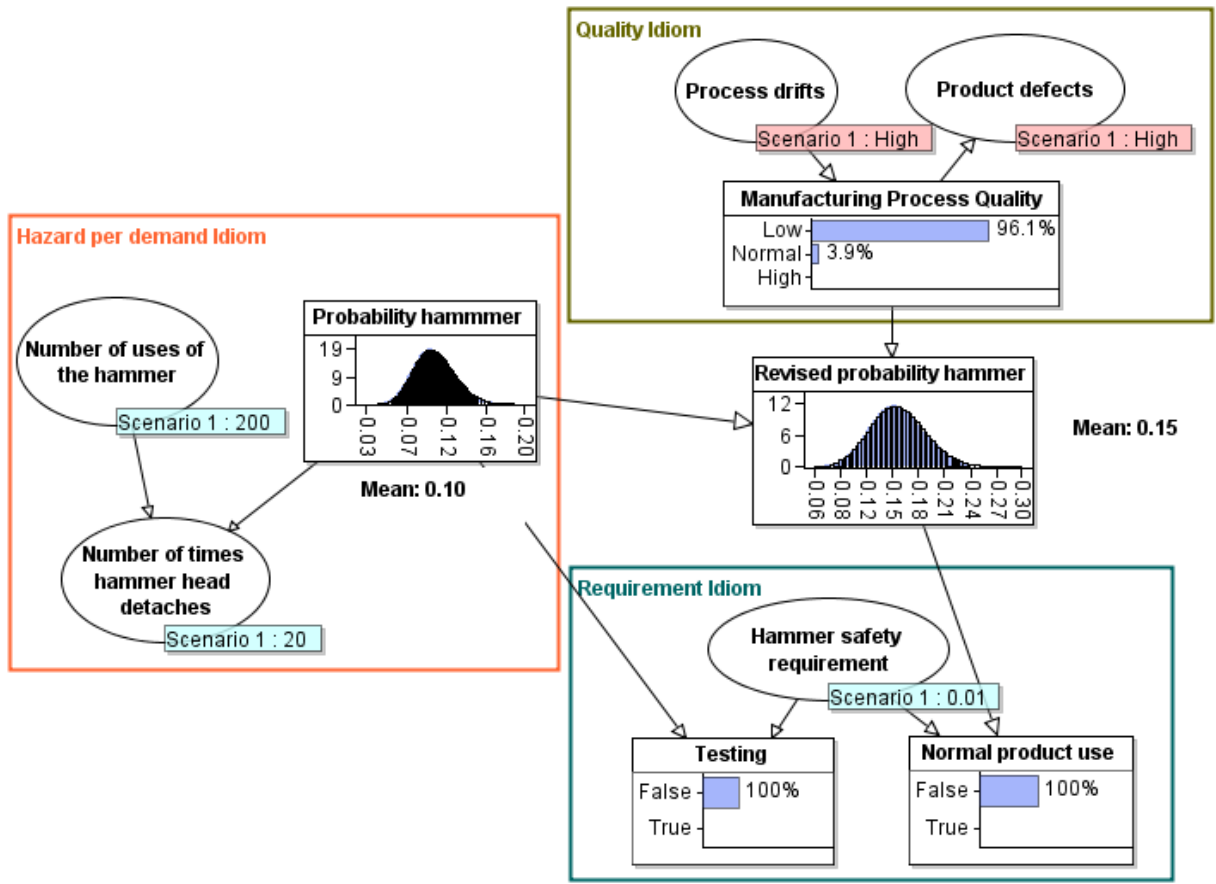

Figure 13. Example of a BN fragment to estimate the reliability of a hammer

### 3.2. Idioms for modelling product failures, hazards and injuries occurrences

Determining the occurrence of failures or hazards and related injuries for a product are essential for informing appropriate risk control measures to prevent harm to users and damage to the environment. In this section, we describe idioms associated with determining the occurrence of failures or hazards and related injuries for a product. These idioms address interaction faults and system degradation that can result in failures or hazards and subsequently harm to the user. There are three idioms in this category:

1. Hazard or failure occurrence idiom
2. Injury event (occurrence) idiom
3. Product injury idiom

### 3.2.1. Hazard or failure occurrence idiom

System degradation and consumer behaviour when using a product, e.g., misuse and frequency of use, can greatly influence the occurrence of failures or hazards for a product. Therefore, it is essential to understand how these factors impact the occurrence of failures or hazards for a product to reduce potential harm to consumers.

The hazard or failure occurrence idiom shown in Figure 14a is an instance of the cause-consequence idiom (Neil et al., 2000) that models the relationship between a hazard(s) or failure(s) and its causal factors. A factor can be any observable attribute or situation that increases or decreases the likelihood or uncertainty of a hazard or failure occurring, such as consumer behaviour. An instance of this idiom is shown in Figure 14b. In Figure 14b, for the hammer example, if the consumer does not use the hammer as intended (minor deviations from intended use), the mean probability of the hammer head detaching per demand increases from 0.15 to 0.18. Product manufacturers and safety regulators may find this idiom useful since it can incorporate all causal factors that affect the occurrence of failures and hazards for a product.

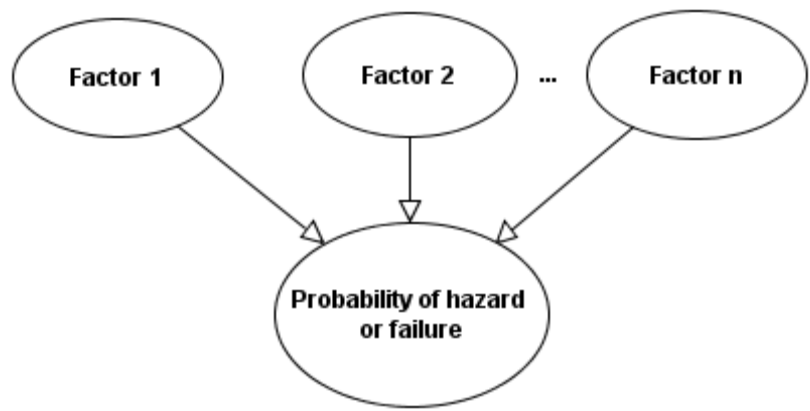

Figure 14a Hazard or failure occurrence idiom

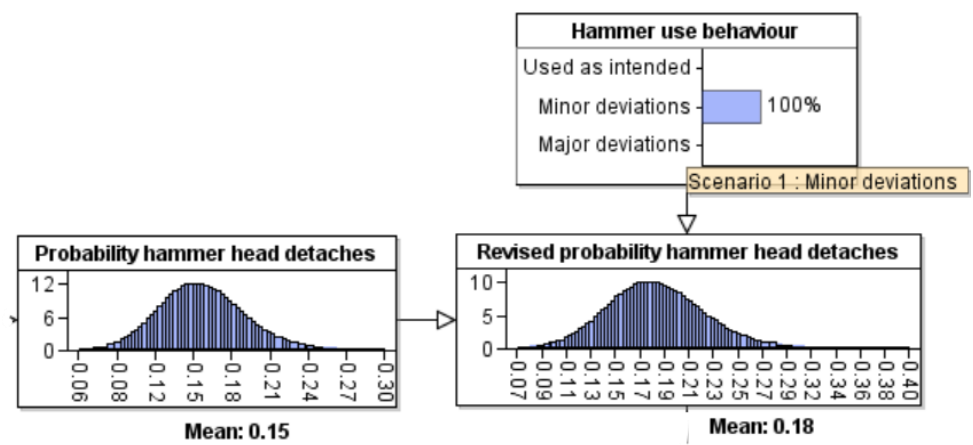

Figure 14b Hazard or failure occurrence idiom instance

### 3.2.2. Injury event (occurrence) idiom

Given the injury scenario for a product, we will be interested in the probability of injury given a failure or hazard. We can estimate the probability of an injury given a failure or hazard by considering the probability of the failure or hazard occurring and the probability of the failure or hazard causing an injury. The probability of the failure or hazard occurring can be estimated using *reliability idioms* (see Section 3.1) and the *hazard or failure occurrence idiom* (see Section 3.2.1); the probability of the failure or hazard causing an injury can be estimated from injury data obtained from reputable sources such as hospitals and injury databases.

The injury event (occurrence) idiom shown in Figure 15a models the probability of an injury event (i.e., an occurrence of injury) during product use. It estimates the probability of an injury event $P(I)$ by combining the probability of the failure or hazard occurring $P(H)$, and the probability of the failure or hazard causing an injury $P(I|H)$ i.e., $P(I) = P(H) \times (I|H)$. An instance of this idiom is shown in Figure 15b. In Figure 15b for the hammer example, if the mean probability of the hammer head detaching and causing a head injury is 0.08 and the mean probability of the hammer head detaching is 0.18, then the estimated mean probability of a head injury occurring while using the hammer is 0.015.

Please note that for the injury event idiom we are assuming a single known type of hazard; however, a product usually has multiple potential hazards. In situations where a product has multiple potential different hazards that are unique in terms of properties they possess, e.g., small parts, electric shock and toxicity, we can add other nodes to the idiom representing different hazards. However, in situations where the hazards, e.g., hot surfaces, open flames and hot gases though unique, are similar in terms of properties they possess, we can identify and define hazard groups or classes, e.g., 'extreme temperature'. The idiom can use the defined hazard groups to consider multiple similar hazards rather than a single hazard.

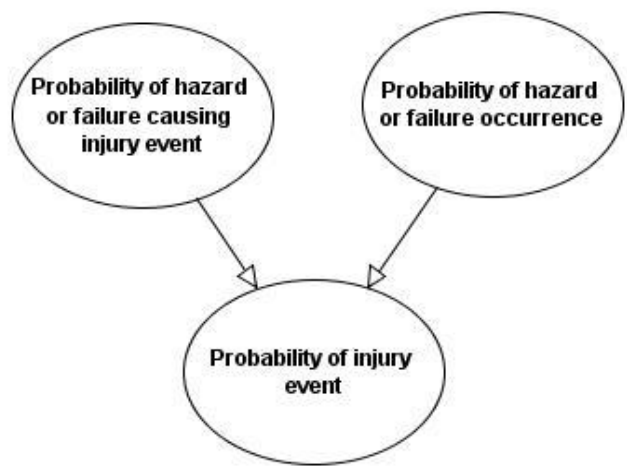

Figure 15a. Injury event idiom

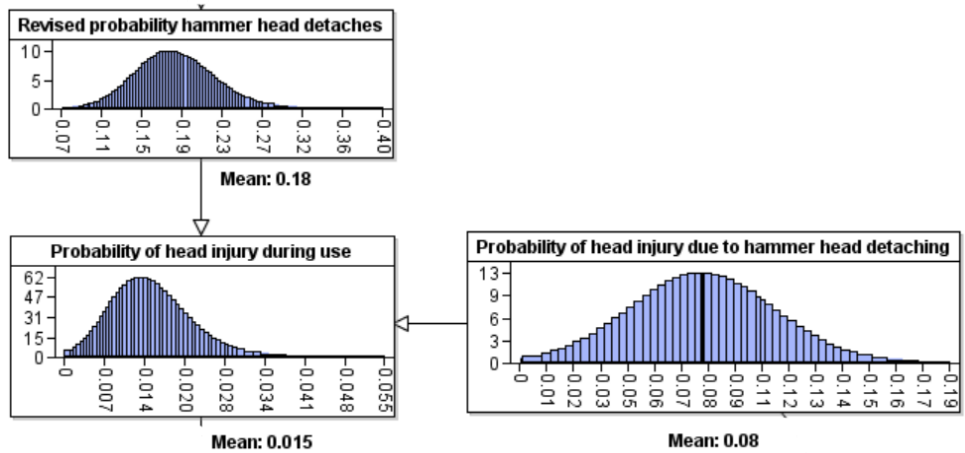

Figure 15b. Injury event idiom instance

### 3.2.3. Product injury idiom

For some products, we may be interested in estimating the number of injuries due to product failures, hazards or hazardous situations. In these situations, we have to consider the probability of the injury event and the number of product instances (i.e., the total number of products manufactured or available on the market). The probability of the injury event can be obtained using the *injury event idiom* (see Section 3.2.2), and the number of product instances can be obtained using manufacturing or sales data.

The product injury idiom shown in Figure 16a models the number of injury events for a set of product instances. This idiom uses a Binomial distribution for the number of injury events. An instance of this idiom is shown in Figure 16b. In Figure 16b, for the hammer example, the idiom estimates that the mean number of head injuries is 1500. In this example, we assume there are 100000 hammer instances, and the mean probability of a head injury is 0.015 (see Figure 15b).

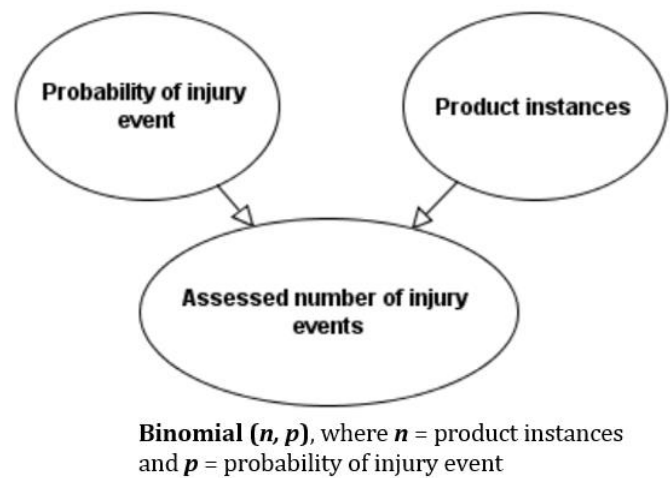

Figure 16a Product injury idiom

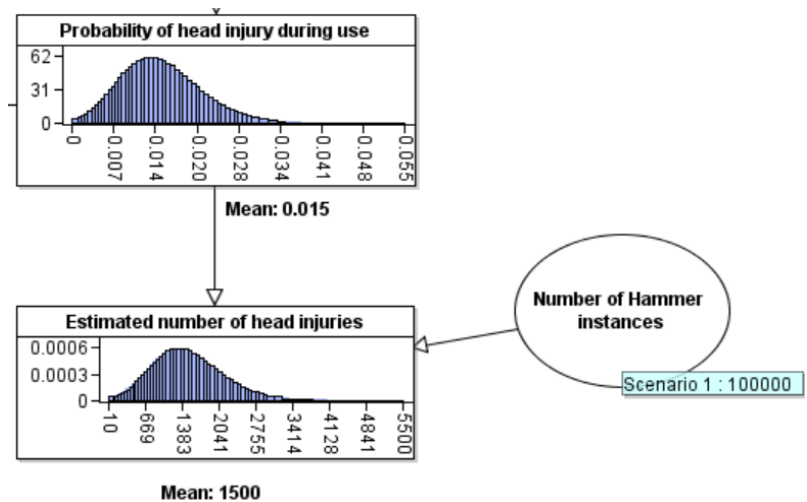

Figure 16b Product injury idiom instance

### 3.3. Idioms for modelling product risk

Determining the overall risk of the product is essential for informing risk management decisions such as product recalls and risk controls. In this section, we describe idioms associated with determining the risk of a product. These idioms satisfy the final task of the risk estimation phase (see Figure 1), i.e., determine the overall risk of the product. There are two idioms in this category:

1. Risk control idiom
2. Risk idiom

### 3.3.1. Risk control idiom

For most products, we may be interested in estimating the effect of risk controls on the occurrence of failures, hazards and related injuries. In these situations, we need to consider the probability of the risk control to mitigate the event (i.e., failures, hazards and injuries) and the probability of the event occurring in the absence of risk controls. *Risk control* is any measure or action taken to mitigate the consequence of an event.

The risk control idiom shown in Figure 17a models the effect of risk controls on an event. It uses the probability of the risk control to mitigate the event $C$, and the probability of the event $E$, to compute the residual probability of the event consequence $RE$ i.e., $RE = (1 - C) \times E$. The risk control idiom can be adapted to model the occurrence of hazards and harm (injury). An instance of this idiom is shown in Figure 17b. In Figure 17b, for the hammer example, the idiom computes that the mean probability of a head injury is 0.04 after risk controls are implemented. In this example, we assume that the mean probability of a head injury in the absence of risk controls is 0.08 and the probability of the risk control mitigating the head injury is 0.5.

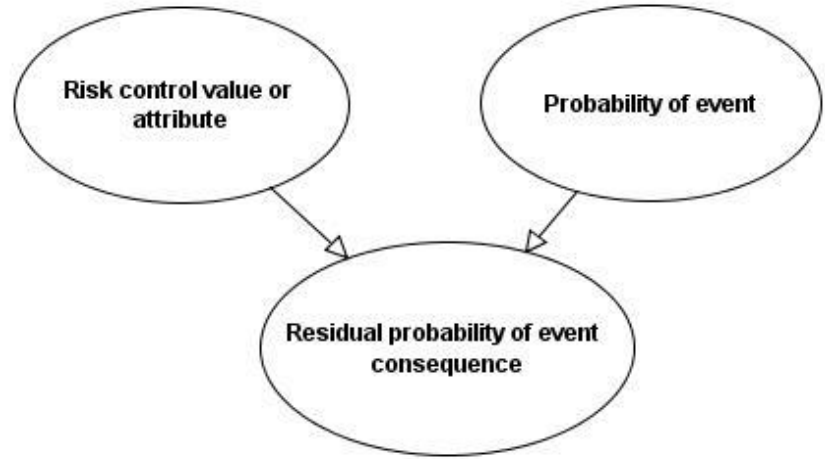

Figure 17a Risk control idiom (generic)

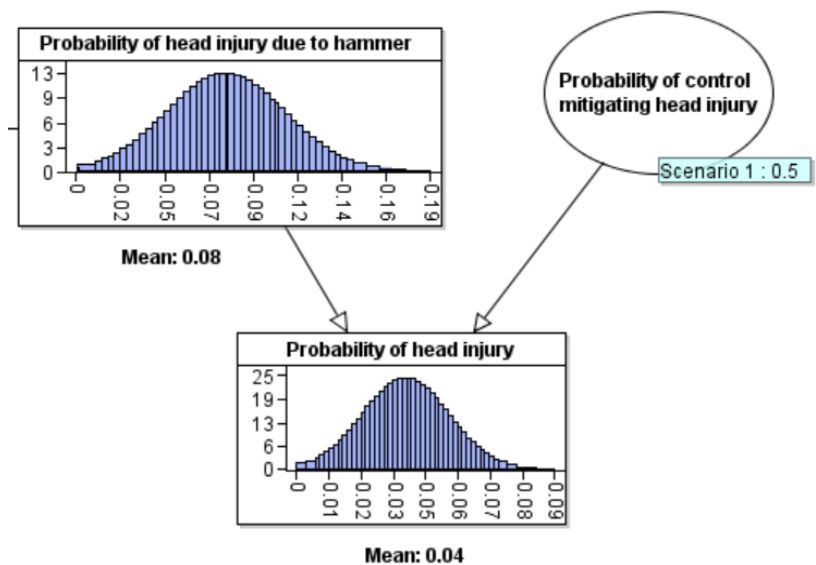

Figure 17b Risk control idiom instance

### 3.3.2. Risk idiom

Previous product safety idioms provide the probability distributions for events, including failures, hazards and injuries associated with a product and its use. We can use this information to estimate the risk of a product using the risk idiom. The risk idiom shown in Figure 18a is used to generate a discrete risk score (e.g., a 5-point scale for regulatory purposes) that is a combination of a set of complex measures. This idiom model risk in terms of its factors and is a special case of the generic definitional idiom (Neil et al., 2000); however, the specific mapping from the continuous function into a discrete set will be specific to the context. For example, in RAPEX, the risk level for a consumer product is defined based on specific injury probability bounds and injury severity levels. For instance, a product is judged as 'low risk' given any injury severity level if the probability of the product causing an injury is less than 0.000001. An instance of the risk idiom is shown in Figure 18b. In Figure 18b, for the hammer example, the idiom estimates the risk of the hammer using a ranked node (Fenton, Neil, &

Caballero, 2007) with a 5-point scale ranging from 'very low' to 'very high' considering the probabilities of the hammer causing a head injury and minor injuries, respectively. In this example, there is a 98% chance that the risk of the hammer is 'very high'.

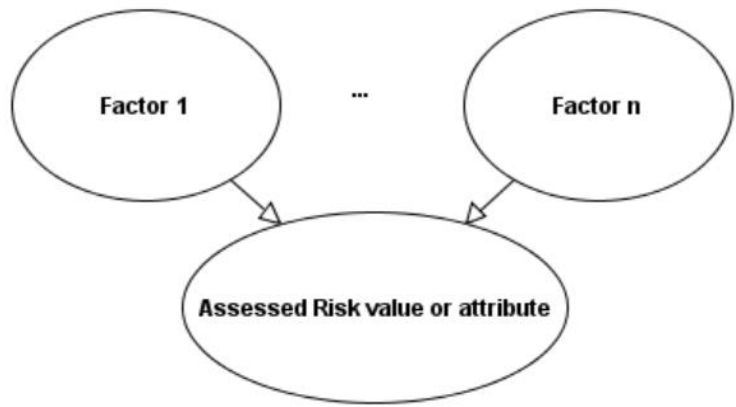

Figure 18a Risk idiom

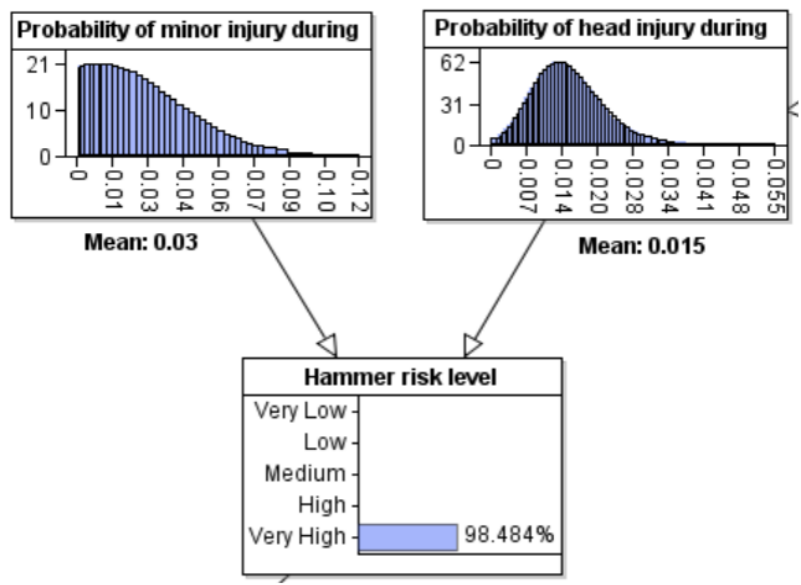

Figure 18b Risk idiom instance

## 4. Risk Evaluation idioms

The last phase of the risk assessment process is risk evaluation (see Figure 1). Risk evaluation "is the process by which the outcome of the risk analysis is combined with policy considerations to characterise the risk and inform decisions on risk management" (Hunte et al., 2022; OPSS, 2021). It entails determining whether the estimated risk of the product is acceptable or tolerable given its benefits. In this section, we describe the two idioms for risk evaluation:

1. Risk tolerability (acceptability) idiom
2. Consumer risk perception idiom

**4.1. Risk tolerability (acceptability) idiom**

In situations where the overall risk of a product is judged unacceptable and additional risk controls are not practical, the product manufacturer or safety regulator may need to determine if the benefit of the product outweighs its risks. The risk tolerability (acceptability) idiom shown in Figure 19a models the trade-off between risk and benefit (or utility) for a product. It evaluates whether the estimated risk score (level) of a product is acceptable or tolerable given the benefit (or utility). The benefits of a product may be determined from literature or consumer surveys. An instance of this idiom is shown in Figure 19b. In Figure 19b, for the hammer example, we define the benefit and risk values using ranked nodes (Fenton, Neil, & Caballero, 2007). Given that the benefit of the hammer is considered average ('medium') and the risk of the hammer is 'very high', then the risk tolerability for the hammer will be 'low' (95% chance the risk tolerability is 'low' or 'very low').

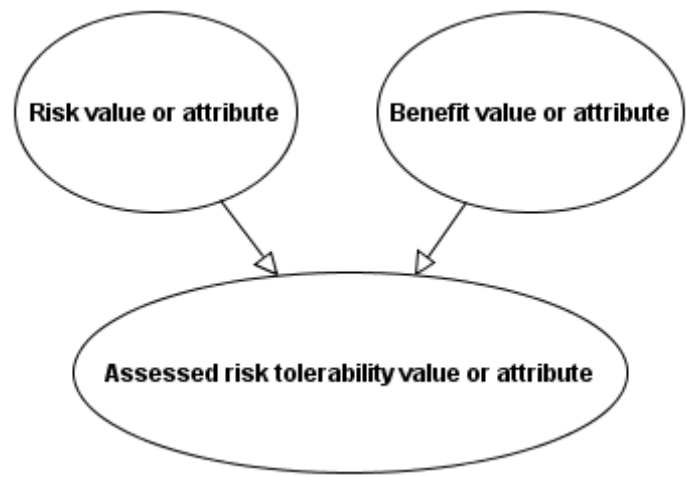

Figure 19a Risk tolerability idiom

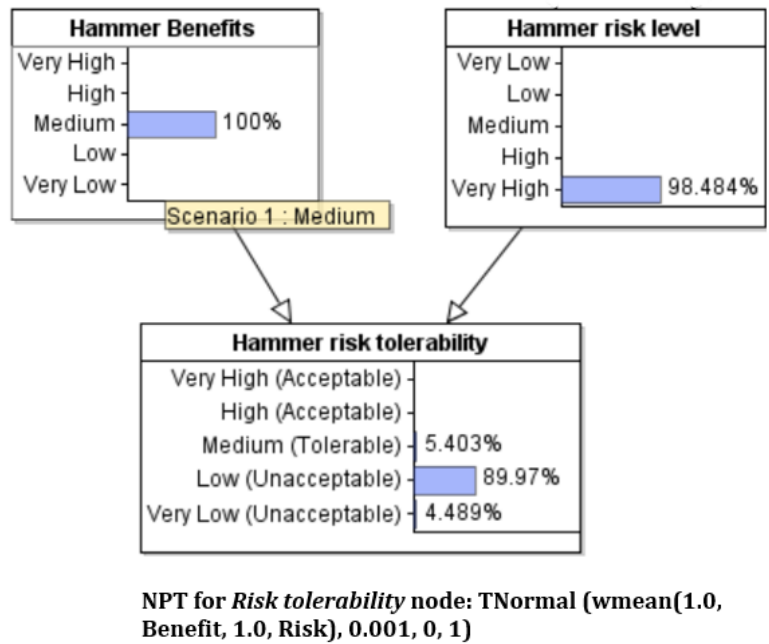

NPT for *Risk tolerability* node: TNormal (wmean(1.0, Benefit, 1.0, Risk), 0.001, 0, 1)

Figure 19b Risk tolerability idiom instance

## 4.2. Consumer risk perception idiom

Consumers may judge the risk and benefits of products differently from experts. For instance, experts tend to judge the risk of a product using quantitative risk assessments, whereas consumers judge risk using a combination of subjective measures such as risk propensity. Therefore, it is essential to understand consumers' perceived risk and benefits of a product to inform risk management decisions. Since the actual value of consumers' perceived risk or benefits may be latent or difficult to measure, we have to use measurable indicators and causal factors to estimate their perceived risk and benefits.

The consumer risk perception idiom shown in Figure 20a estimates consumer risk perception of a product using causal factors (or interventions) and indicators. Please note that this idiom does not incorporate different user profiles. An instance of this idiom is shown in Figure 20b and Figure 20c. In Figures 20b and 20c, for the hammer example, we define the variables using ranked nodes (Fenton, Neil, & Caballero, 2007). In Figure 20b, the idiom shows that consumers may perceive the risk of the hammer as 'high' since they judge the likelihood of injury and severity of the injury as 'high'. In Figure 20c, the idiom shows the impact of a product recall, negative media stories and consumer feedback on consumer risk perception of the hammer.

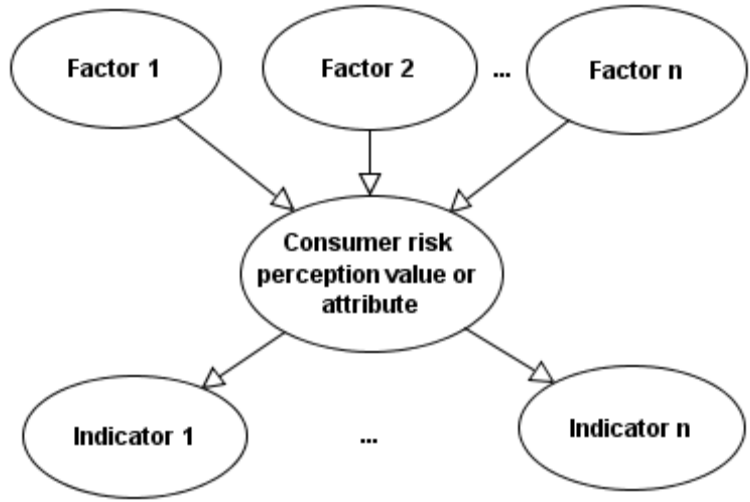

Figure 20a Consumer risk perception idiom

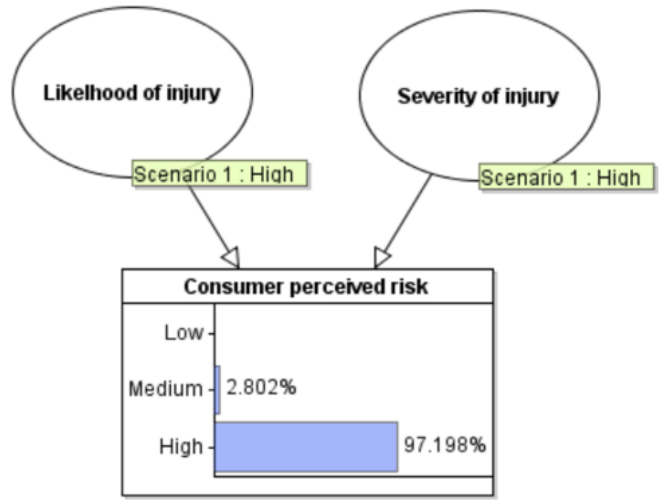

NPT for *Consumer perceived risk* node:
TNormal (wmean(1.0, injurylikelihood, 1.0, injuryseverity), 0.001, 0, 1)

Figure 20b Consumer risk perception idiom instance 1

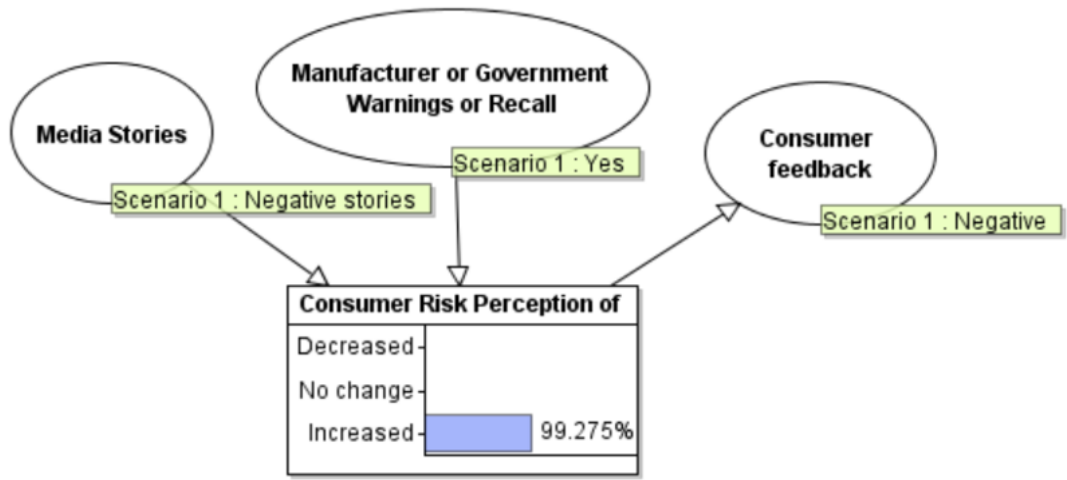

Figure 20c Consumer risk perception idiom instance 2

## 5. Putting it all together: Consumer products and aircrafts BN examples

The product safety idioms have been used to build BNs for different product safety cases. In this section, we discuss BNs used to assess the safety and risk of consumer products (see Section 5.1) and aircraft reliability (see Section 5.2).

### 5.1. Example 1: Consumer product safety and risk assessment BN

The generic BN model for consumer product safety and risk assessment shown in Figure 21 was developed by Hunte, Neil & Fenton (2022) to assess the risk of consumer products using relevant information about the product, its users and its manufacturer. Examples of the product safety idioms used to develop the BN are highlighted in Figure 21. The proposed BN model resolved the limitations with traditional methods like RAPEX and has shown good predictive performance, especially in situations where there is little or no product testing data. For further details on the BN model, such as node probability tables and case study results, please see Hunte et al. (2022).

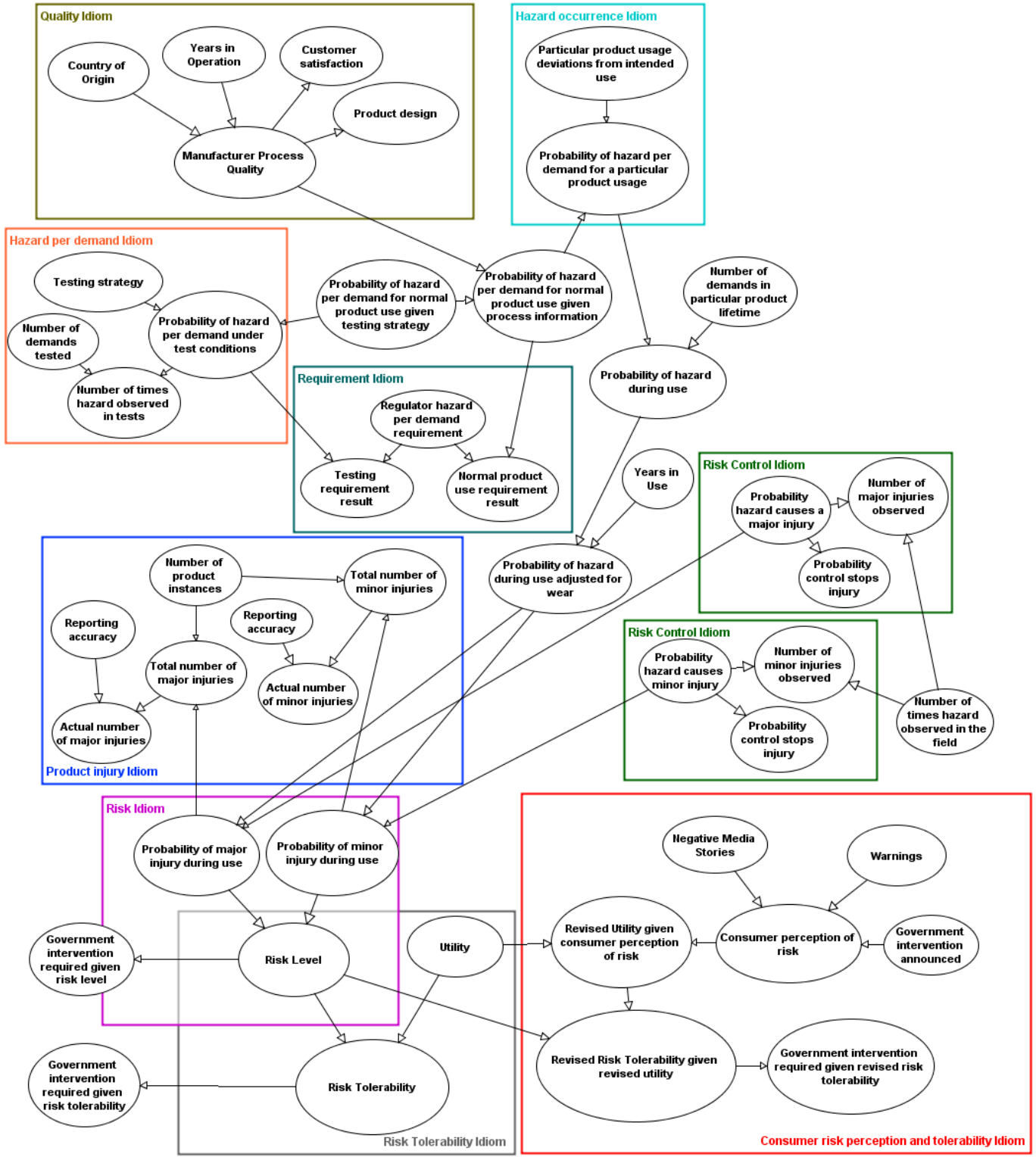

Figure 21. Product risk assessment BN model developed by Hunte et al. (2022) with visible product safety idioms

## 5.2. Example 2: Aircraft reliability BN

The aircraft reliability BN shown in Figure 22a shows a fragment of the safety assessment for a new military aircraft that focuses on estimating the probability of failure during a mission due to engine and/or braking system failure. It incorporates both TTF and PFD measures to determine overall reliability since the reliability measure for the engine is TTF, and the braking system is PFD. The product safety idioms connected causally to estimate the reliability of a military aircraft during a mission are highlighted in Figure 22a.

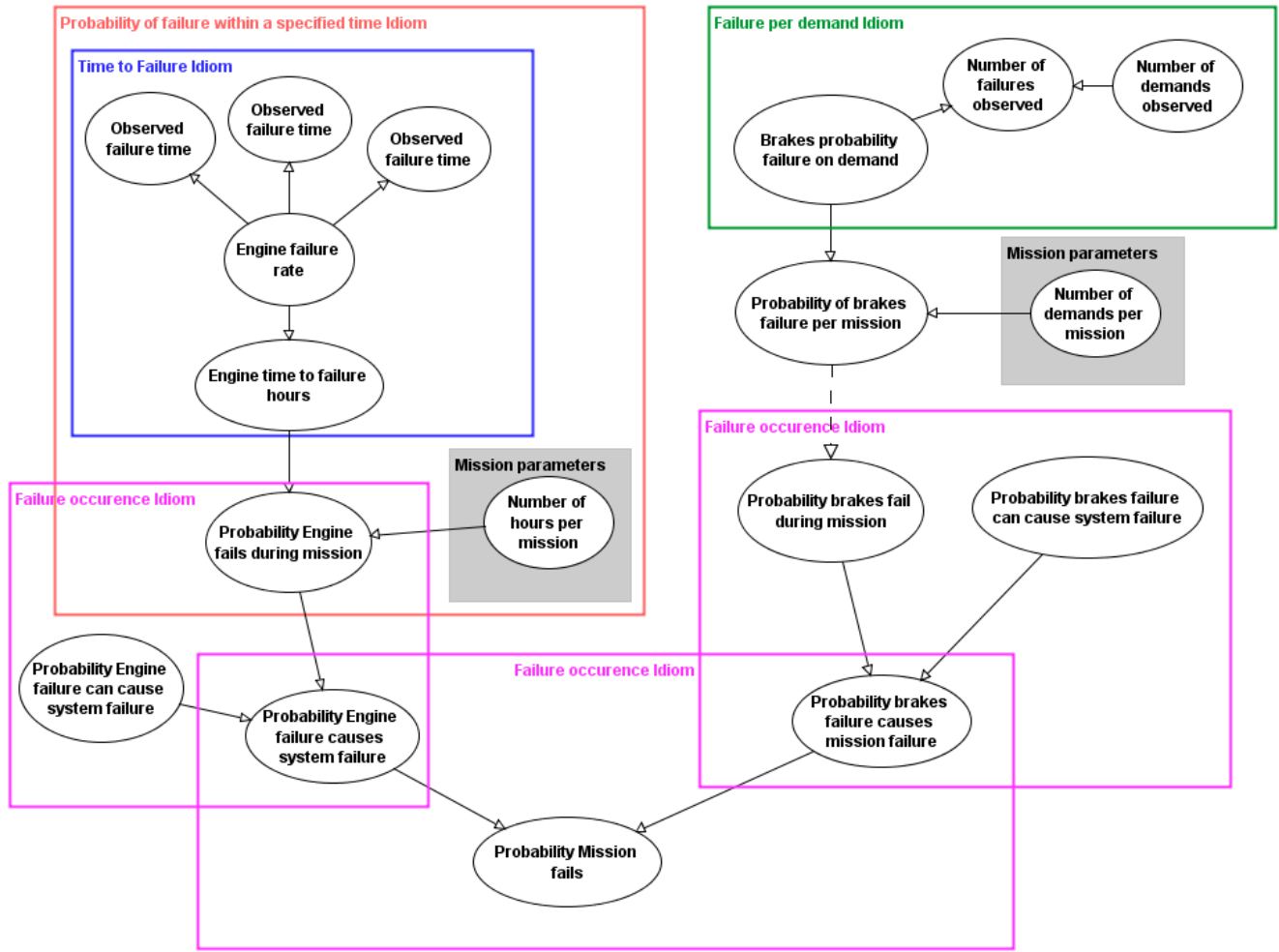

Figure 22a. Aircraft reliability BN with visible product safety idioms

In Figure 22b, the BN model estimates the probability of failure for a military aircraft during a mission due to engine failure and braking system failure is 0.0008 (0.08%). In this example, we assume that for the engine, we observed failure times of 6000, 5000 and 4000 hours, respectively and the engine is used for 6 hours during the mission. We assume that there is a 50% chance that the engine can cause a system failure. For the braking system, we assume that we observed 10 failures in 1000000 demands and the braking system is used once during the mission. We also assume that there is a 50% chance that the braking system can fail. Please note that this BN model can be extended to incorporate other aircraft systems such as flight control systems to determine the overall reliability of an aircraft.

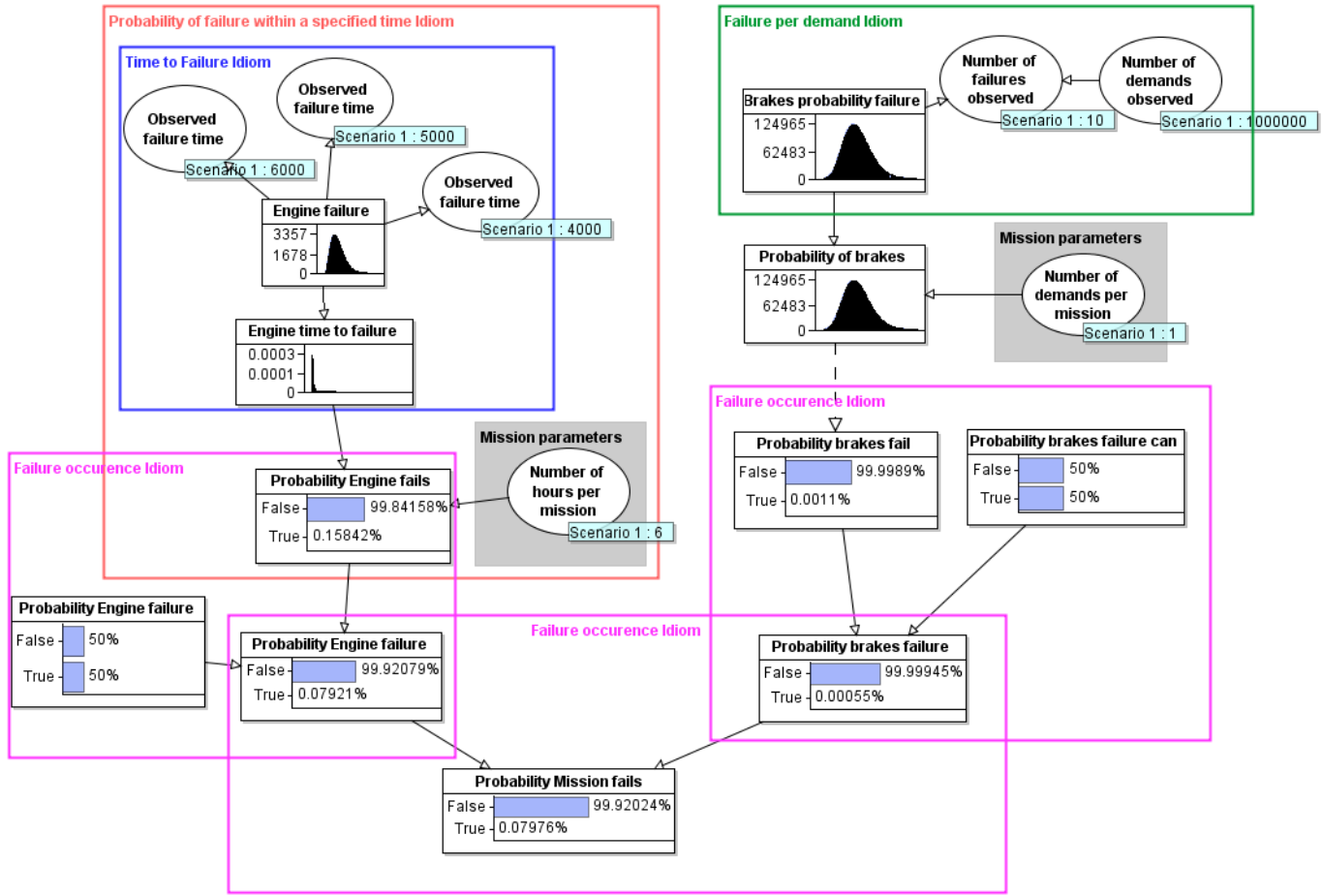

Figure 22b. Aircraft reliability BN instance

## 6. Conclusion and recommendation

There is no established method for eliciting variables and building BNs specifically for product safety and risk assessment. This paper introduces a novel set of idioms, called *product safety idioms,* to enable a systematic method for developing BNs specifically for product safety and risk assessment. While the proposed idioms are sufficiently generic to be applied to a wide range of product safety cases, they are not prescriptive or complete and should be considered as a guide for developing suitable idioms for product safety and risk assessments given available product-related information. Product manufacturers, safety regulators and safety and risk professionals will find the proposed product safety idioms useful since they cover the main activities of product safety assessments, i.e., risk estimation and risk evaluation, and offer the following benefits:

1. *Handles limited and incomplete data*: The idioms can combine objective and subjective evidence to provide reasonable risk estimates for products, especially in situations where there is limited or no historical testing and operational data.

2. *Standardise product safety BN development*: The idioms provide a library of reusable BN patterns for product safety that facilitates the easy development of practical product safety BNs. They also guide the knowledge elicitation process by allowing risk assessors to identify relevant information (known or unknown) required to build custom idioms and BNs for product safety assessments.

3. *Enhance communication, interpretability and explainability*: The structure and results of product safety BNs developed using the idioms can be easily interpreted, explained, and reviewed by risk assessors and safety regulators. Risk assessors can easily justify the structure and results of the BN. Also, the product safety idioms can serve as a validation method for future product safety and risk BNs ensuring that their structure is practical and logical.

We believe that the product safety idioms discussed in this paper are meaningful reasoning patterns that guide the development of complex BNs for product safety and risk assessments. Future work includes applying the idioms to many different product safety cases.

**Acknowledgements**

This work was supported by the UK Government Department for Business, Energy and Industrial Strategy, Office for Product Safety and Standards (OPSS) and Agena Ltd. The views expressed in this article are those of the authors exclusively, and not necessarily those of the UK Government Department for Business, Energy and Industrial Strategy, Office for Product Safety and Standards (OPSS).

**Declaration of Interest/ Conflict of Interest**

Norman Fenton and Martin Neil are Directors of Agena Ltd.